\documentclass[]{elsarticle}
\usepackage[utf8]{inputenc}
\usepackage{amsmath}
\usepackage{graphicx}
\usepackage{float}
\usepackage{booktabs}
\usepackage{algorithm}
\usepackage[noend]{algpseudocode}
\usepackage{catchfile}
\usepackage{pdflscape}
\usepackage{longtable}

\begin{document}

\begin{frontmatter}

\title{Probabilistic Grammars for Equation Discovery}
\author[1,2]{Jure Brence\corref{cor1}}
\ead{jure.brence@ijs.si}
\author[3,1]{Ljupčo Todorovski}
\ead{ljupco.todorovski@fu.uni-lj.si}
\author[1,2]{Sašo Džeroski}
\ead{saso.dzeroski@ijs.si}

\address[1]{Jožef Stefan Institute, Department of Knowledge Technologies, Jamova cesta 39, 1000 Ljubljana, Slovenia}
\address[2]{Jožef Stefan International Postgraduate School, Jamova cesta 39, 1000 Ljubljana, Slovenia}
\address[3]{University of Ljubljana, Faculty of Public Administration, Gosarjeva ulica 5, 1000 Ljubljana, Slovenia}

\cortext[cor1]{Corresponding author}


\begin{abstract}
Equation discovery, also known as symbolic regression, is a type of automated modeling that discovers scientific laws, expressed in the form of equations, from observed data and expert knowledge. Deterministic grammars, such as context-free grammars, have been used to limit the search spaces in equation discovery by providing hard constraints that specify which equations to consider and which not. In this paper, we propose the use of probabilistic context-free grammars in equation discovery. Such grammars encode soft constraints, specifying a prior probability distribution on the space of possible equations.
We show that probabilistic grammars can be used to elegantly and flexibly formulate the parsimony principle, that favors simpler equations, through probabilities attached to the rules in the grammars. We demonstrate that the use of probabilistic, rather than deterministic grammars, in the context of a Monte-Carlo algorithm for grammar-based equation discovery, leads to more efficient equation discovery. Finally, by specifying prior probability distributions over equation spaces, the foundations are laid for Bayesian approaches to equation discovery.
\end{abstract}


\end{frontmatter}

\section{Introduction}

Scientific data are ever-more readily available in large amounts for use in building models in the form of equations by scientists from different disciplines. This has recently triggered a sharp increase in  the interest in equation discovery, also known as symbolic regression.
Equation discovery aims at automated discovery of quantitative laws, expressed in the form of equations, in collections of measured data~\cite{dzeroski2007,schmidt2009}. History of science offers plenty of instances of the equation discovery task: given a collection of measured data, find quantitative laws, expressed in the form of equations. Take for example, Newton's second law of motion that relates the acceleration $ a $ of an observed moving object, its mass $ m $ and the force applied to it, $ F $, in the well known equation $ F = m \cdot a $. As part of an equation discovery task, the collected measured data includes measurements of $ m $, $ a $ and $ F $.

The task of equation discovery is closely related to the task of supervised regression. Machine learning methods for supervised regression assume a fixed class of models, e.g., linear or piece-wise linear, and find the one that provides the best fit to the training data. Equation discovery methods typically consider broader classes of mathematical equations. These classes may be vast and many (often infinitely many) equations can be found that provide excellent fit to the training data. The challenge of symbolic regression is therefore twofold. On one hand, one can easily overfit the training data with an unnecessarily complex equation. On the other hand, the space of candidate equations is huge and grows exponentially as equation complexity increases, posing serious computational issues to equation discovery methods.

Based on the way they address these challenges, equation discovery methods can be clustered into two general paradigms. The first paradigm deploys a general and powerful search algorithm to the unconstrained space of candidate equations. Most commonly, symbolic regression methods employ genetic algorithms~\cite{schmidt2009,gomes2019}. Recently, alternative approaches based on sparse regression~\cite{kutz} or neural networks~\cite{tegmark} have been proposed. The second paradigm, commonly referred to as equation discovery, focuses on the use of knowledge for constraining the space of candidate equations. The different types of knowledge being considered include measurement units~\cite{sds}, grammars~\cite{lagramge, sebag}, cross-domain knowledge ~\cite{pret} and domain-specific knowledge~\cite{pbm, scipm}.

A notable limitation of the existing methods following the equation-discovery paradigm is that they can only use \textsl{hard} constraints, i.e., they can only include/exclude a certain equation from the list of candidates. In this paper, we extend the equation discovery paradigm and propose probabilistic context free grammars (PCFGs) as a formalism for constraining the space of candidate equations. In addition to hard constraints, probabilistic grammars allow for specifying \textsl{soft} constraints that specify preferences over the space of candidates by assigning a prior probability to each of them.

The conjecture of this paper is that the use of probabilistic context free grammars holds promise for a significant impact on the computational efficiency of equation discovery methods. To test the validity of the conjecture, we consider one hundred equations from the Feynman data set~\cite{tegmark}. For each of them, we observe the expected number of candidate equations that have to be considered for different approaches to reconstruct the correct equation. We perform a theoretical comparison of the expected number of candidates for a deterministic grammar and different variants of a probabilistic context-free grammar for arithmetic expressions. We also analyze the impact of the prior distribution over the space of candidate equations on the reconstruction performance of a simple Monte-Carlo sampling algorithm for equation discovery\footnote{Our code is publicly available at https://github.com/brencej/ProGED.}.

The remainder of the paper is organized as follows. Section~2 introduces probabilistic context-free grammars, in general, and their specific instances for arithmetic expressions that we are going to use in our analysis, in particular. In Section~3, we perform a theoretical analysis of the expected number of candidate equations that have to be considered for reconstructing the Feynman equations. Section~4 reports the results of the empirical comparison of the reconstruction performance of different sampling procedures for equation discovery. Section~5 discusses the analysis results and puts them in the context of related work on symbolic regression and equation discovery. Section~6 provides a summary of our work and a number of directions for further research.

\section{Grammars for specifying structural priors}
This section introduces the notion of a grammar. Originating from computational linguistics, grammars are used as formal specifications of languages and use a set of production rules to derive valid strings in the language of interest. Our paper argues for the use of context-free grammars, a subclass of general grammars that is general enough to specify the language of arithmetic expressions. In equation discovery, we are interested in using grammars as generative models, as opposed to their common use for parsing, i.e., discriminating between legal and illegal strings in a language. In the rest of this section, we first introduce context-free grammars and their probabilistic counterparts through formal definitions and illustrative examples. We then define a PCFG for arithmetic expressions and define the task of (probabilistic) grammar-guided equation discovery.


\subsection{Context-free grammars}
Formally, a context-free grammar \cite{ConciseIntroductionLanguages} is defined as a tuple $G=(\mathcal{N},\mathcal{T},\mathcal{R},S)$. The set $\mathcal{T}$ contains terminal symbols, i.e., words for composing sentences in the language. Non-terminal symbols in the set $\mathcal{N}$ represent higher-order terms in the language, such as noun or verb phrases. The production rules in the set $\mathcal{R}$ are rewrite rules of the form $A \rightarrow \alpha$, where the left-hand side is a non-terminal, $A \in \mathcal{N}$, while the right-hand side is a string of non-terminals and terminals, $\alpha \in (\mathcal{N} \cup \mathcal{T})^{*}$. In natural language, a rule $\mathit{NP} \rightarrow {A} \, {N}$ specifies that a noun phrase $\mathit{NP}$ is a sequence of an adjective $A$ and a noun $N$. These two non-terminals represent the subsets of terminals corresponding to adjectives and nouns, respectively. When deriving a sentence, a grammar starts with a string containing a single non-terminal $S \in \mathcal{N}$ and recursively applies production rules to replace non-terminals in the current string with the strings on the right-hands sides of the rules. The final string, containing only terminal symbols, belongs to the language defined by the grammar $G$.

Consider a simple example of a context-free grammar $G_L = (\mathcal{N}_L, \mathcal{T}_L, \mathcal{R}_L, S_L)$ deriving linear expressions from two variables $x$ and $y$:
\begin{enumerate}
    \item $\mathcal{N}_L = \{E,V\}$,
    \item $\mathcal{T}_L = \{x,y,+\}$,
    \item $\mathcal{R}_L = $\begin{equation}\label{eq:grammar_linear}
    \begin{aligned}
   E &\rightarrow E + V \\
   E &\rightarrow V \\
   V &\rightarrow x \\
   V &\rightarrow y,
   \end{aligned}
   \end{equation}
   \item $S_L = E$.
\end{enumerate}
We usually write multiple production rules with the same non-terminal on the left-hand side using a compact, single-line representation, $E \rightarrow E + V \mid V$.

\begin{figure}[h]
    \centering
    \includegraphics[width=0.45\textwidth]{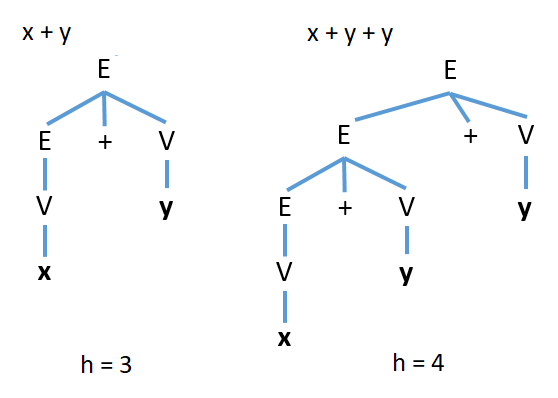}
    \caption{Two parse trees, derived by the grammar $G_L$, defined in Equation~\ref{eq:grammar_linear}. Left: $\psi^L_1 = \psi^L(``x+y")$, right: $\psi^L_2 = \psi^L(``x+y+y")$. \label{im:parse-trees}}
\end{figure}

The process of deriving a string in the language of a context-free grammar is depicted with a parse tree. A rooted, labeled and ordered tree $\psi$ is a parse tree, derived by grammar $G$, if its root is labeled $S$, the labels of the leaf nodes are terminal symbols from $\mathcal{T}$, and the immediate children of an internal node labeled $A \in \mathcal{N}$ correspond to a left-to-right sequence of the symbols in $\alpha$, where $A \rightarrow \alpha \in \mathcal{R}$ \cite{ConciseIntroductionLanguages}. The string derived by $\psi$ is the left-to-right sequence of its leafs. Figure~\ref{im:parse-trees} provides two example parse trees for the grammar $G_L$ above, $\psi^L_1$ and $\psi^L_2$, for the expressions $x+y$ and $x+y+y$, respectively.
We measure the height of the parse tree as the number of edges on the longest path from the root node ($S$) to a leaf. The heights of parse trees $\psi^L_1$ and $\psi^L_2$ are three and four, respectively. The same mathematical expression will in general have a different parse tree (and height) when derived by using a different grammar. Nevertheless, the height of a parse tree is closely related to the complexity of the derived mathematical expression, with more complex expressions requiring higher parse trees.

\subsection{Probabilistic context-free grammars}
A grammar can be turned into a probabilistic grammar by assigning probabilities to each of its productions, such that for each $A \in {\cal N}$:
\begin{equation*}
    \label{eq:PCFG-productions}
    \sum\limits_{(A \rightarrow \alpha) \in \mathcal{R}}P(A \rightarrow \alpha) = 1.
\end{equation*}
The probabilities of all productions with the same nonterminal symbol on the left hand side sum up to one. In other words, we impose a probability distribution on the productions with the same symbol on the left hand side. As each parse tree $\psi$, derived by a grammar $G$, is characterized by a sequence of productions, its probability is simply the product of the probabilities of all productions associated with it \cite{geman}:
\begin{equation}
    \label{eq:parse-tree-probability}
    P(\psi) = \prod\limits_{(A \rightarrow \alpha)\in \mathcal{R}} P(A \rightarrow \alpha)^{f(A \rightarrow \alpha, \psi)},
\end{equation}
where $f(A \rightarrow \alpha, \psi)$ denotes a function counting the number of instances of the production $A\rightarrow \alpha$ in the parse tree $\psi$.
Furthermore, it can be shown that for any proper PCFG the probabilities of all parse trees derived by the grammar sum up to one~\cite{chi}:
\begin{equation}
    \label{eq:parse-tree-probability-sum}
    \sum\limits_{\psi \in \Psi}P(\psi) = 1.
\end{equation}
We can extend the example context-free grammar from Equation \ref{eq:grammar_linear} to a probabilistic context-free grammar by assigning a probability to each of the four productions, given below in brackets after each production:
\begin{equation}
    \label{eq:grammar-linear-pcfg}
    \begin{aligned}
        E &\rightarrow E + V \ [p] \mid V \ [1-p] \\
        V &\rightarrow x \ [q] \mid y \ [1-q],
    \end{aligned}
\end{equation}
where we have parametrized the probability distributions over productions for $E$ and $V$ the with parameters $0 < p < 1$ and $0 < q < 1$, respectively. We can now calculate the probabilities of the two example parse trees $\psi^L_1 = \psi^L("x+y")$ and $\psi^L_2=\psi^L("x+y+y")$ from Figure~1:
\begin{equation}
    \label{eq:example-probability}
    \begin{aligned}
        P(\psi^L_1) &= p(1-p)q(1-q),\\
        P(\psi^L_2) &= p^2(1-p)q(1-q)^2.
    \end{aligned}
\end{equation}
Longer and more complex mathematical expressions are composed by using more productions and their associated parse trees have thus lower probabilities.

Throughout the paper, where the distinction between grammars and their probabilistic counterparts is important, we refer to the former as deterministic (context-free) grammars.

\subsection{Grammars as generators}

While grammars were invented to formally describe the structure of natural-language sentences, in computer science, they are mainly used for specifying the syntactic structure of programming and markup languages~\cite{ConciseIntroductionLanguages}. In this context, grammars allow for the implementation of efficient parsers of source code and annotated documents: For a given string and a grammar, the parser determines whether the string can be derived using the grammar and, if yes, return the appropriate parse tree. The latter explicates the syntactic structure of the string.

In equation discovery, context-free grammars have been used as generators of arithmetic expressions~\cite{lagramge}. The deterministic algorithm for generating expressions from a given grammar performs systematic enumeration of parse trees from simpler (lower) to more complex (higher) ones. To this end, the algorithm employs a refinement operator that for a given parse tree, generates its minimal refinements by replacing simpler production rules with more complex ones \cite{lagramge}. The deterministic generator requires a user-specified maximal height of the generated parse trees. Note that, in such a setting, each generated parse tree is assumed to be equally probable, i.e., the uniform distribution over the space of parse trees with a limited height is assumed.

\begin{algorithm}[t!]
\caption{Sample a sentence from a probabilistic context-free grammar.}
\begin{algorithmic}[1]
  \Require{Probabilistic grammar $G=(\mathcal{N},\mathcal{T},\mathcal{R},S)$, non-terminal $A \in \mathcal{N}$}
  \Ensure{Sentence $s$ corresponding to a randomly sampled parse tree $\psi$ from $G$ with root node $A$, probability $p$ of $\psi$}
  \Procedure{generate\_sample}{$G$, $A$}
  \State $(s, p)$ = ([\ ], 1)
  \State Choose a random rule $(A \rightarrow \alpha) \in \mathcal{R}: \alpha = A_1 A_2 \ldots A_k, \, A_i \in \mathcal{N} \cup \mathcal{T}$
  \For{$i=1$, $i \leq k$}
  \If{$A_i \in \mathcal{T}$}
  \State $s$ = $s$.append($A_i$)
  \Else
  \State $(s_i, p_i)$ = \textsc{generate\_sample}($G$, $A_i$)
  \State $s$ = $s$.append($s_i$)
  \State $p$ = $p \cdot p_i$
  \EndIf
  \EndFor
  \State \textbf{return} $(s, p)$
  \EndProcedure
\end{algorithmic}
\label{alg:generate_sample}
\end{algorithm}

Probabilistic grammars, on the other hand, provide a much more flexible generation mechanism, based on sampling the space of parse trees that can be derived by a grammar, described in Algorithm~\ref{alg:generate_sample}. The function \textsc{generate\_sample} takes two arguments: the probabilistic context-free grammar $G$ and a non-terminal $A$ as the root node of the parse tree of the generated expression. The procedure recursively expands its current list of elements $s$ until it contains only terminal symbols. In each recursive step, it randomly samples production rules for the non-terminal $A$, according to their probability distribution as prescribed by the grammar (line~3). The sequence of symbols on the right-hand side of the chosen production rule is then used to expand the current sentence $s$: the extension is simple for terminals (line~6) and includes recursion for non-terminals (lines~8 to 10). When called with a grammar and its start symbol, the algorithm returns a randomly sampled sentence from the grammar along with its probability.

In contrast to the deterministic systematic generator of parse trees from a given grammar, Algorithm~\ref{alg:generate_sample} does not need a specification of a maximal tree height. Also, each generated sentence (tree) is assigned a probability, based on the probabilities of the production rules used in its derivation. Therefore, the distribution over the generated parse trees is not necessarily uniform. As we will show in the next section, the more complex (higher) parse trees are less probable than the simpler ones.

\subsection{Number and probabilities of parse trees with limited height}

For a given deterministic grammar $G=({\cal N}, {\cal T}, {\cal R}, S)$, the number of parse trees $n_G(A,h)$ with a height of exactly $h$ and the root symbol $A \in {\cal N} \cup {\cal T}$ can be computed using a recursive formula:
\begin{equation}
    \label{eq:counting-n}
    n_G (A, h) = 
    \begin{cases}
    1 & \text{if $A \in \mathcal{T},\ h=0$} \\
    0 & \text{if $A \in \mathcal{T},\ h>0$} \\
    0 & \text{if $A \in \mathcal{N},\ h=0$} \\
    \text{number of productions: $A \rightarrow w, \ w \in \mathcal{T}^{*}$} & \text{if $A \in \mathcal{N},\ h=1$} \\
    \sum\limits_{(A \rightarrow \alpha) \in \mathcal{R}} \left( \prod\limits_{i=1}^{k} N_G(A_i, h-1) - \prod\limits_{i=1}^{k} N_G(A_i, h-2) \right) & 
    \text{if $A\in \mathcal{N},\ h>0$}.
    \end{cases}
\end{equation}
\noindent
Here, $\alpha = A_1 A_2 \ldots A_k$ is a string of symbols $A_i \in {\cal N} \cup {\cal T}$, while $N_G(A, h)$ represents the number of parse (sub)trees of $G$ with a root node $A$ and height up to (and including) $h$:
\begin{equation*}
    \label{eq:counting-N}
    N_G(A,h) = \sum\limits_{h_i=0}^h n_G(A, h_i).
\end{equation*}
To compute the number of parse trees derived by a grammar, we use the above formula with $A=S$. Whenever the root symbol is not specified, we assume $n_G(h) = n_G(A=S, h)$. We can now, due to the simplicity of the linear grammar $G_L$, use the above formula to find an analytical expression for the number of parse trees at a given height:
$n_{G_L}(h) = n_V ^ {h-1}, \, \forall h \geq 2$. Details are provided in the Appendix.

Equation~\ref{eq:counting-n} introduces a recursive approach to counting the number of parse trees with a particular height derived by a grammar. The number of parse trees $N_G(A=S, h)$ does not change if we extend a context-free grammar to a probabilistic context-free grammar. However, we can introduce another related quantity that helps characterize the properties of a PCFG, called coverage. We define it as the sum of probabilities (Equation~\ref{eq:parse-tree-probability}) of all the parse trees with height up to (and including) $h$:
\begin{equation}
    \label{eq:def-coverage}
    \text{Cov}_G(A, h) = \sum\limits_{h_i=0}^h \;\; \sum\limits_{\psi \in \Psi_{A,h_i}} P(\psi),
\end{equation}
where $\Psi_{A,h_i} \subseteq \Psi $ denotes the set of all parse trees with the root symbol $A$ and height $h_i$. For a proper grammar, we can now express Equation~\ref{eq:parse-tree-probability-sum} in terms of coverage as
\begin{equation}
    \label{eq:coverage-sum}
    \lim\limits_{h\rightarrow \infty}\text{Cov}_G(S, h) = 1.
\end{equation}
We can compute the coverage of a PCFG at height $h$ using a similar set of recursive equations as in Equation \ref{eq:counting-n}:
\begin{equation}
    \label{eq:counting-Cov}
    \text{Cov}_G(A, h) = 
    \begin{cases}
    1 & \text{if $A \in \mathcal{T},\ h\ge 0$}\\
    0 & \text{if $A \in \mathcal{N},\ h=0$}\\
    \sum_{(A \rightarrow \alpha) \in {\cal R}} P(A \rightarrow \alpha) \prod_{i=1}^{k} \text{Cov}_G(A_i, h-1) & 
    \text{if $A \in \mathcal{N},\ h>0$},
    \end{cases}
\end{equation}
where $\alpha = A_1 A_2 \ldots A_k$ is the string of symbols $A_i \in {\cal N} \cup {\cal T}$ on the right-hand side of the production rule $A \rightarrow \alpha$.

To illustrate the benefit of introducing the probabilities of the production rules, let us reconsider the simple linear grammar $G_L$ from Equation~\ref{eq:grammar_linear}. There are two probability distributions for the production rules we have to assign to $G_L$: one for the production rules with $E$ on the left-hand side and the other for the ones with $V$ on the left-hand side. The latter is trivial, since one would assume that all the variables are equally probable, leading to a uniform distribution of probabilities of the production rules $V \rightarrow v$, i.e., $P(V \rightarrow v) = 1 / n_V$, where $n_V$ denotes the number of variables ($n_V = 2$ in the example grammar from Equation~\ref{eq:grammar_linear}).

The other probability distribution is over the two productions $E \rightarrow E+V$ and $E \rightarrow V$. If we denote the probability of the first (recursive) rule with $p$, the probability of the other rule is $1-p$. Following the recursive formula for coverage from Equation~\ref{eq:counting-Cov}, one can easily show that 
\begin{equation}
    \label{eq:coverage-grammar-linear}
    \text{Cov}_{G_L}(E, h) = 1 - p^{h-1}, \quad \forall h \geq 2.
\end{equation}


\begin{figure}[h!]
    \includegraphics[width=1\textwidth]{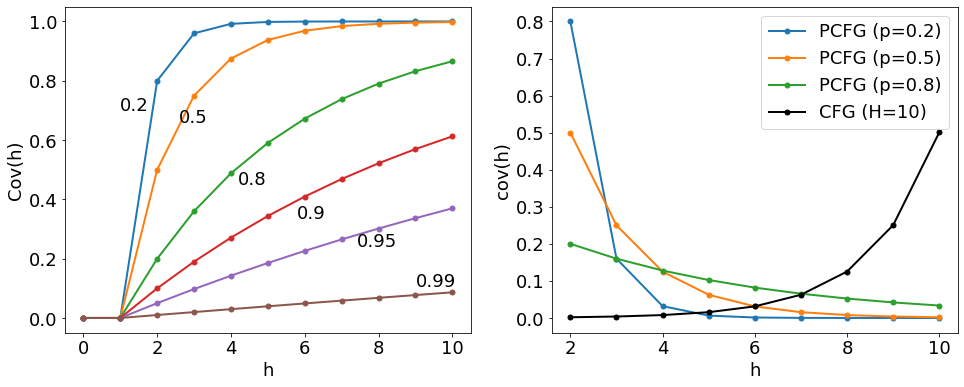}
    \caption{The impact of the probability $p$ of the recursive rule $E \rightarrow E + V$  in $G_L$ on the probability ($\text{cov}(h)$) of all trees with exactly a given height $h$ and coverage ($\text{Cov}(h)$), the probability of all trees up to, and including, a given height $h$. Left: the change of coverage of the probabilistic $G_L$ with increasing $h$ for different values of $p$ (coloured lines). Right: change of the probability to sample a parse tree with increasing height $h$ for probabilistic $G_L$ with changing $p$ (colours) and deterministic $G_L$ (black line).\label{im:coverage}}
    \centering
\end{figure}

This result is important, since it reveals the impact of the probability $p$ of the recursive production rule $ E \rightarrow E + V$ on the probabilities of sampling a parse tree with a certain height, depicted in Figure~\ref{im:coverage}. The probability of sampling the simplest/lowest parse trees at height 2 equals $1 - p$. For small values of $p$, the probability of sampling one of the simplest trees is high. These simplest trees correspond to the expressions $x$ and $y$, i.e., to expressions including a single linear term. Increasing $p$ increases the likelihood of sampling higher parse trees corresponding to more complex expressions. Thus, $p$ allows us a direct and intuitive control over the parsimony principle in equation discovery that introduces preference towards simpler expressions. Some equation discovery systems introduce regularization \cite{kutz} with the same purpose. The regularization constant, however, usually has an unlimited range and does not provide an intuitive, probabilistic interpretation of the parameter controlling the preference for simpler expressions.

Furthermore, the probability distribution of the parse trees in $G_L$ with a certain height is uniform, following the uniform distribution over the production rules for $V$. Since the number of parse trees with height $h$ increases exponentially with $h$, the probability to sample a single tree with a given height $h$ decreases exponentially with $h$. This is in contrast with the behavior of deterministic grammars, where the sampling probability is uniformly distributed across all trees. Since the number of trees increases exponentially with increasing height, it is much more probable to sample a higher tree than a lower one. The use of deterministic grammars for equation discovery requires an external (regularization) mechanism to implement the parsimony principle \cite{lagramge}.

\subsection{PCFGs for arithmetic expressions}
Context-free grammars are typically used to parse sentences. Probabilistic context-free grammars enhance this use, for instance, by providing an estimate of the probability of a sentence in addition to its parse tree. However, probabilistic context-free grammars allow for another type of application - stochastic generation of sentences or, in our case, mathematical expressions, as described in Algorithm~\ref{alg:generate_sample}.
The probabilities, assigned to the productions, provide a great amount of control over the probability distribution of individual parse trees. In our example in Equation \ref{eq:grammar-linear-pcfg}, the parameter $p$ controls the probability of a higher number of terms, while the parameter $q$ tunes the ratio between the number of occurrences of variables $x$ and $y$. 


An important concept to consider when working with grammars is ambiguity. A grammar is formally ambiguous if there exist sentences that can be described by more than one parse tree, generated by the grammar. Grammars for arithmetic expressions can express another type of ambiguity, called semantic ambiguity. All but the simplest arithmetic expressions can be written in many mathematically equivalent, but grammatically distinct ways. This is mainly due to the distributivity, associativity and commutativity properties of the basic operations. For example, consider the many ways the expression $x^2 + xy$ can be written: $x\times x + x\times y = x\times y + x\times x = x \times (x + y) = x \times (y + x) = (x + y)\times x = (y + x)\times x$. Each results in a distinct parse tree.
It is generally useful to adopt a canonical representation that each generated equation is converted into. This allows us to compare expressions to each other and check whether they are mathematically equivalent in addition to comparing their parse trees. In our work, we use the Python symbolic mathematics library SymPy \cite{sympy} to simplify expressions and convert them into a canonical form, as well as to compare expressions symbolically.

For equation discovery, we are going to use a formally unambiguous context-free grammar for arithmetic expressions:
\begin{enumerate}
    \item $\mathcal{N} = \{E, F, T, V \}$
    \item $\mathcal{T} = \{x, y, +, -, *, /, (, ) \}$
    \item $\mathcal{R} = $ \begin{equation}
        \label{eq:grammar-universal}
        \begin{aligned}
            E &\rightarrow E + F \ \mid \ E - F \ \mid \ F \\
            F &\rightarrow F * T \ \mid \ F / T \ \mid \ T \\
            T &\rightarrow (E) \ \mid \ V \\
            V &\rightarrow x \ \mid \ y.
        \end{aligned}
    \end{equation}
    \item $S = E$.
\end{enumerate}

In this paper, we illustrate a number of useful properties of PCFG for equation discovery on the example of the universal arithmetic grammar. The grammar in Equation \ref{eq:grammar-universal} can be extended to a PCFG by imposing a probability distribution to each set of production rules with the same nonterminal on the left hand side. In addition, we include four mathematical functions into the grammar: $\sin{}$, $\cos{}$, $\sqrt{}$ and $\exp{}$. The resulting PCFG is defined in Equation \ref{eq:grammar-universal-pcfg}:
\begin{enumerate}
    \item $\mathcal{N} = \{E, F, T, R, V \}$
    \item $\mathcal{T} = \{c, x, y, +, -, *, /, (, ), \sin{}, \cos{}, \sqrt{}, \exp{} \}$
    \item $\mathcal{R} = $ \begin{equation}
        \label{eq:grammar-universal-pcfg}
        \begin{aligned}
            E &\rightarrow E + F \ [0.2] \ \mid \ E - F \ [0.2] \ \mid \ F \ [0.6] \\
            F &\rightarrow F * T \ [0.2] \ \mid \ F / T \ [0.2] \ \mid \ T \ [0.6] \\
            T &\rightarrow R \ [0.2] \ \mid \ V \ [0.4] \ \mid \ c \ [0.4] \\
            R &\rightarrow ( E ) \ [0.6] \ \mid \ \sin(E) \ [0.1] \ \mid \ \cos(E) \ [0.1] \mid \sqrt{E} \ [0.1] \ \mid \ \exp(E) \ [0.1] \\ 
            V &\rightarrow x \ [0.5] \ \mid \ y \ [0.5]
        \end{aligned}
    \end{equation}
    \item $S = E$.
\end{enumerate}
We treat the production probabilities in the PCFG as parameters of the grammar that define the prior distribution over expressions and shape the search space of equations for equation discovery. The specific probabilities above are our default values, a choice based on intuition and preliminary experiments.
In the above definition, we have added another new concept to the basic version in Equation \ref{eq:grammar-universal} - the constant parameter $c$. In this paper we discuss PCFG properties in the context of generating expressions for the task of equation discovery. Broadly speaking, there are three components to an arithmetic expression in the context of equation discovery: variables, constants and operators. Numerical values and constants are typically treated as free parameters to be optimized when evaluating an equation for its fit against the given data. We thus include the terminal symbol $c$ in the PCFG to stand for a constant parameter of an expression with as yet-undetermined value.

\subsection{Grammar-guided equation discovery}

After introducing probabilistic context-free grammars for arithmetic expressions, we can now define the task of grammar-guided equation discovery as follows, as an extension of the task considered by Todorovski and Džeroski in 1995~\cite{lagramge}:
\begin{description}
\item{\textsl{Given}}
\begin{itemize}
\item $D$, a data set with observations of a set of variables $V$,
\item $v \in V$, the variable on the left-hand side of the equations,
\item $G=(\mathcal{N}, \mathcal{T}, \mathcal{R}, S)$, a grammar generating arithmetic expressions involving the variables from $V$ except the variable $v$, i.e., $V \setminus \{v\} \subseteq \mathcal{T}$.
\end{itemize}
\item{\textsl{Find}} an equation $v = e$, such that
\begin{itemize}
\item $e$ can be derived using $G$,
\item $e$ minimizes the discrepancy between the observed values of $v$ from $D$ and the values of $v$ calculated by using the equation.
\end{itemize}
\end{description}

The measure of discrepancy used in this paper is the relative root mean square error, i.e.,
\begin{equation}
\label{eq:rermse}
\textit{ReRMSE}(v=e, D) = \frac{1} {\sigma_v} \sqrt{\frac{1}{|D|} {\sum_{o \in D} {(v_o - e_o)^2}}},
\end{equation}
where $|D|$ denotes the number of observations in the data set $D$, $v_o$ and $e_o$ are the values of $v$ and $e$ for a specific observation $o \in D$, respectively, and $\sigma_v^2$ denotes the variance of variable $v$ in the data.





\section{Theoretical analysis}
After introducing context-free grammars and illustrating their use for generating arithmetic expression, we turn our attention towards evaluating the benefits of using PCFGs for equation discovery. We first formulate the task of equation discovery based on PCFGs. Then, we analyze the expected number of parse trees we have to sample from a given (deterministic or probabilistic) grammar in order to reconstruct a specific equation. Finally, we evaluate the impact of the probabilities of production rules on the expected number of sampled trees.

The analysis is based on the one hundred equations from the Feynman Symbolic Regression Database \cite{tegmark}. This database includes a diverse sample of equations from the three-volume set of physics textbooks by Richard P. Feynman \cite{feynman1} and has been previously used as a benchmark for equation discovery \cite{tegmark}. The equations are listed in the appendix and contain between one and nine variables, the four basic operations, the functions $\exp$, $\sqrt{}$, $\sin$, $\cos$, $\tanh$, $\arcsin$ and $\log$, as well as a variety of constants--mostly rational, but also $e$ and $\pi$.

\subsection{Expected number of equations}
The metric used for analysis in this section is the expected number of parse trees that we need to sample from a given grammar, so that the sample includes a parse tree corresponding to a given (target) equation. The value of the metric depends on the target and the grammar. We are going to model the number of sampled parse trees as a random variable $N$, being interested in its expected value $E[N]$.

We model the sampling of parse trees from a given PCFG as a Bernoulli process of consecutive trials. In each trial, we sample a parse tree from the PCFG and check whether it matches the target equation. If we denote the probability of the parse tree corresponding to the target equation with $p$, then $N$ follows the geometric distribution:
\begin{equation}
\label{eq:geometric_distribution}
P(N = n) = (1-p)^{n-1}p.
\end{equation}
\noindent In other words, the Bernoulli process includes a sequence of $n-1$ unsuccessful trials, where the sampled parse tree does not correspond to the target (hence, the probability of each trial outcome is $1-p$), followed by a single successful trial with the probability of $p$. In the Appendix, we derive the formula for $E[N]$, i.e., the expected number of trials until (and including) the first success:
\begin{equation}
    \label{eq:EN}
    E_{PCFG}[N]=\frac{1}{p},
\end{equation}
\noindent which corresponds to the general intuition that the more probable the target equation is, the fewer samples are needed to reconstruct it. 


When dealing with a deterministic grammar, we derive the expected number of sampled trees as follows. First, let us denote the height of the parse tree corresponding to the target equation with $h$. Note however, that the generator does not have the exact value of $h$ at input. Thus, it would have to systematically generate all the parse trees with heights at most $h-1$ to find out that the tree corresponding to the target equation is not among them. Then, the sampling proceeds at height $h$, where the generator will end up identifying the target tree. On average, due to the uniform distribution of parse trees, half of the parse trees with height $h$  will be considered in the last iteration, making the total number of trees considered by a deterministic grammar $G$
\begin{equation}\label{eq:expectedN_deterministic}
E_{CFG}[N] = N_G(h-1) + \frac{1}{2}n_G(h).
\end{equation}

To calculate the probability $p$ and the height $h$ for a specific target equation, we parse the expression on its right-hand side with the grammar of interest.


\subsection{Probabilistic vs deterministic}
Using the formulas derived in the previous subsection, we calculate the expected number of sampled parse trees necessary to reconstruct each of the one hundred equations from the Feynman database. Then, for different values of $n$, we are interested in how many of the target equations from the Feynman database we can expect to reconstruct by taking at most $n$ sample parse trees from a given grammar, i.e.,
\begin{equation}
\label{eq:cumulative_portion}
\text{ratio}(n) = \frac{1}{100}\sum\limits_{i}^{100}I(n \ge E[N_i]),
\end{equation}
where $n$ denotes the number of sampled parse trees, $E[N_i]$ is the expected number of sampled trees necessary to reconstruct the $i$-th equation from the Feynman database, and $I(b)$ is a function indicating the truthfulness of the Boolean expression $b$, having the value of 1, if $b$ is true, and 0 otherwise.

\begin{figure}[t!]
    \centering
    \includegraphics[width=0.95\textwidth]{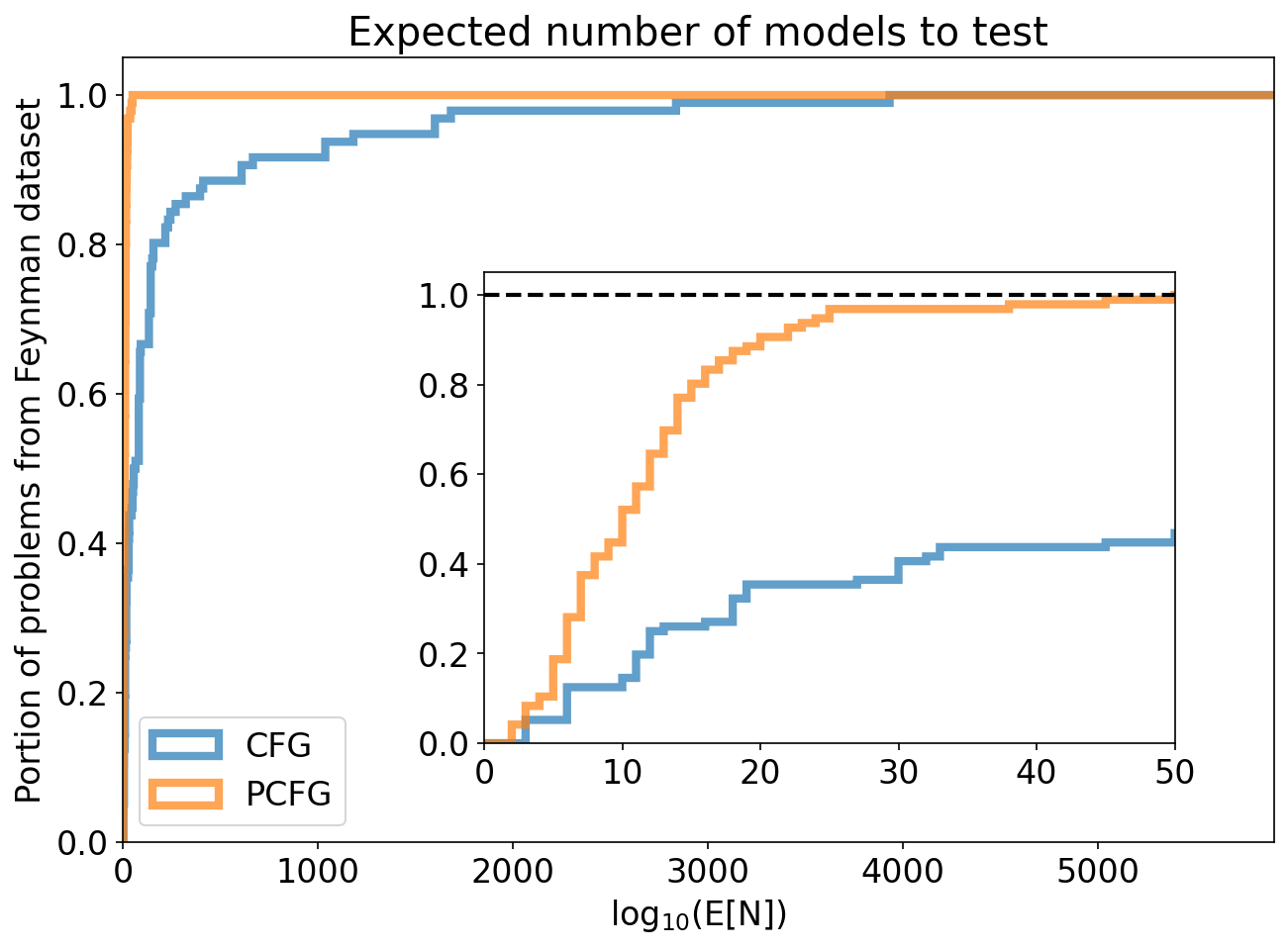}
	\caption{The expected portion of the one hundred Feynman equations (on the y-axis) that can be reconstructed by sampling an expected number of parse trees (on the x-axis) from the probabilistic (PCFG, red line) and the deterministic grammar (CFG, blue line) from Equations~\ref{eq:grammar-universal-pcfg}~and~\ref{eq:grammar-universal}, respectively. The inset provides a zoom-in on the range of the expected number of sampled parse trees below $10^{50}$.\label{im:CFG-vs-PCFG}}
\end{figure}

We compare these ratios for the deterministic and probabilistic grammars for generating arithmetic expressions from  Equations~\ref{eq:grammar-universal}~and~\ref{eq:grammar-universal-pcfg}, respectively. The comparison, depicted in Figure~\ref{im:CFG-vs-PCFG}, shows that
the expected number of samples for the deterministic grammar is many orders of magnitude above the expected number for the probabilistic grammar. In order to reconstruct all the equations in the Feynman database, we need to sample over $10^{4000}$ parse trees from the deterministic grammar. In contrast, the probabilistic grammar would require, on average, less than $10^{50}$ samples. Half of the Feynman equations can be successfully reconstructed when sampling around $10^{10}$ parse trees from the probabilistic grammar, while roughly $10^{60}$ samples are required for the same achievement using the deterministic version of the grammar.

The number of distinct parse trees for a context-free grammar increases super-exponentially with parse tree height. This is true for both probabilistic and deterministic grammars. Yet, the differences in the expected number of sampled expressions observed here span tens, even hundreds of orders of magnitude. This is due to the fact that the probability distribution over parse trees generated by a deterministic grammar is uniform. In contrast, probabilistic grammars introduce bias towards simpler parse trees, as we have shown in Section 2. The results in Figure~\ref{im:CFG-vs-PCFG} show that the bias introduced by probabilistic grammars corresponds well to the equations included in the Feynman database, indicating that the bias corresponds well to equations used in science. In the next section, we will show how a careful adjustment of the probabilities assigned to individual production rules in the grammar can further shift the bias towards equations in the Feynman database.

\subsection{Biased vs. unbiased probabilistic grammar}
Here, we focus our attention on probabilistic grammars, and explore the impact of tuning the prior distribution over the parse trees, via changing the probabilities of production rules, on the expected number of sampled trees. Recall that the analysis in Figure~\ref{im:coverage} shows that decreasing the probabilities of the recursive production rules introduces bias towards simpler equations. We now extend that analysis by exploring the impact of the probabilities of the other production rules on the expected number of sampled trees. To keep the comparative analysis simple and clear, we vary the following four parameters:
\begin{enumerate}
    \item the ratio between the probabilities of summation and subtraction:
    
    $r_{\text{sum}} = \frac{P(E \rightarrow E+F)}{P(E \rightarrow E-F)},$
    \item the ratio between the probabilities of multiplication and division:
    
    $r_{\text{mul}} = \frac{P(F \rightarrow F*T)}{P(F \rightarrow F/T)},$
    \item the ratio between the probabilities of a constant and a variable:
    
    $r_{\text{const}} = \frac{P(F \rightarrow V)}{P(F \rightarrow c)},$
    \item the ratio of the total probability of the set of the four special functions and no special function:
    
    $r_{\text{funct}} = \frac{P(R \rightarrow \sin(E)) + P(R \rightarrow \cos(E)) + P(R \rightarrow \sqrt{E}) + P(R \rightarrow \exp(E))}{P(R \rightarrow (E))}.$
\end{enumerate}
\noindent The first grammar in our analysis sets all of these ratios to one.  We will refer to this grammar as the uniform grammar, since it does not introduce any preferences among the operators, functions or type of atomic terms (variables or constants) in the equation. In contrast, the biased grammar introduces intuitive preferences often used by scientists and engineers.
Table~\ref{tab:parameters} reports the exact parameter settings for the two grammars.

\begin{figure}
    \centering
    \includegraphics[width=1\textwidth]{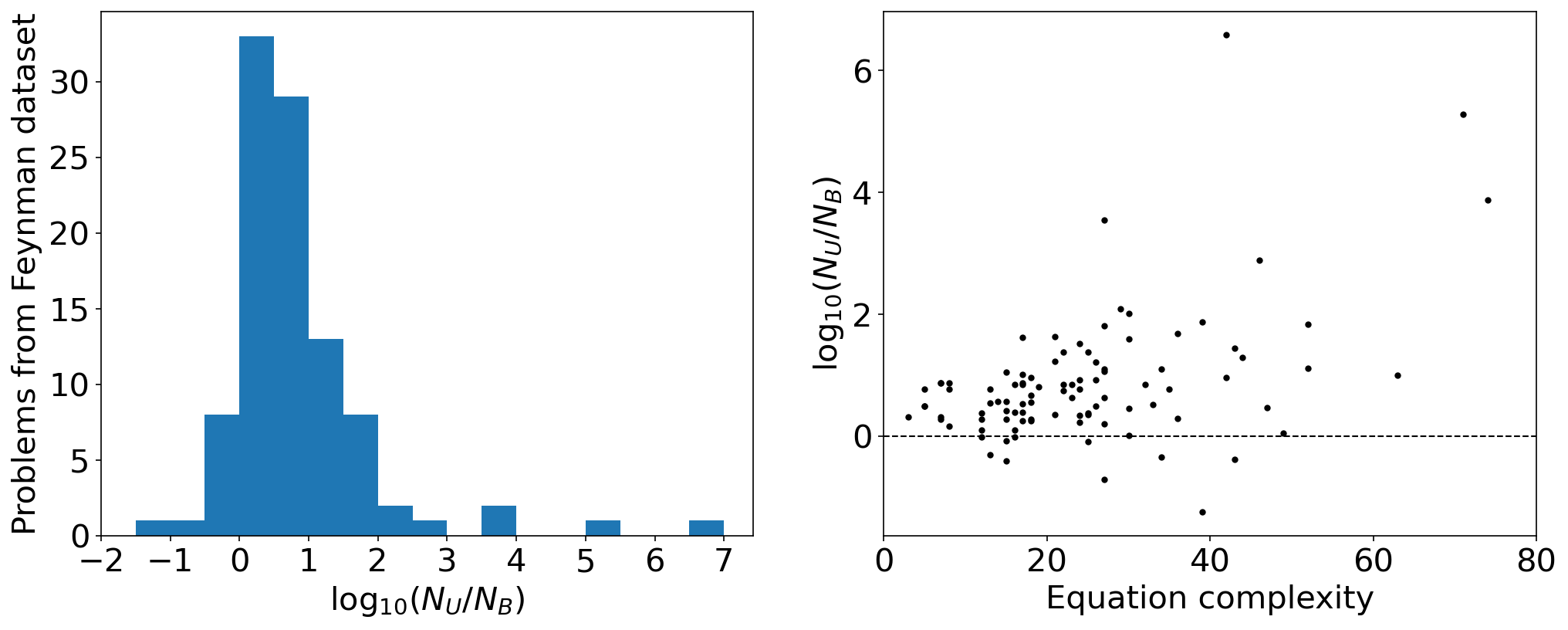}
    \caption{Reduction of the expected number of sampled parse trees needed to reconstruct the hundred target equations from the Feynman database when introducing bias in the probabilistic grammar for arithmetic expressions. Left: histogram depicting the number of Feynman equations (y-axis) with a certain reduction value (x-axis). Right-hand side: scatter plot of the reduction factor (y-axis) and equation complexity (x-axis) for each Feynman equation. $N_U$ and $N_B$ indicate the expected number of sampled parse trees for the uniform and the biased universal grammar, respectively.\label{im:speedup}}
\end{figure}

\begin{table}[ht]
       \setlength{\tabcolsep}{0.8em}
	   \centering
       \caption{Parameter values for the uniform and the biased universal grammars.}
	   \begin{tabular}{@{}lcccr@{}}
	        \\
	        \toprule
	       	  & $r_{\text{sum}}$ & $r_{\text{mul}}$  & $r_{\text{const}}$ & $r_{\text{funct}}$   \\ \midrule
    	    \multicolumn{1}{l}{\textbf{Uniform}} &  1 & 1 & 1 & 1      \\
	       	\multicolumn{1}{l}{\textbf{Biased}} & 0.4 & 1.5 & 0.25 & 0.67          \\ \bottomrule
	       \end{tabular}
        \label{tab:parameters}
\end{table}
The results of the comparison are depicted in Figure~\ref{im:speedup}. The histogram on the left-hand side of the figure shows that introducing the bias into the probabilistic grammar leads to a nontrivial reduction in the number of expected samples, albeit smaller than the one observed when comparing the probabilistic and the deterministic grammar. The biased grammar reduces the expected number of samples for 86 out of the one hundred Feynman equations, whereas an increase is observed for merely ten equations (for four equations, the expected number of samples is the same for both grammars). Of the 86 cases with reduction, 58 lie in the range of a $25\%$ reduction in the expected number of parse trees at worst, and a $90\%$ reduction at best. For 28 target equations, the expected number of models is reduced by more than one order of magnitude, where the largest reduction is on the scale of seven orders of magnitude. 

The biased grammar seems to exploit the deeper aspects of the basic parsimony principle. One might expect then, that the biased grammar would lead to a greater reduction in the number of expected samples for simpler target equations. Note however, that the scatter plot on the right-hand side of Figure~\ref{im:speedup} shows the opposite trend: the reduction tends to increase with the increasing complexity of the equation, with the Spearman correlation coefficient between the reduction rate and complexity being 0.5. In this particular figure, we measured the complexity of the target equation as the length (the number of alphanumerical characters) of its string/text representation. Table~A.6, in the appendix, summarizes the results of an analysis which also considers alternative measures of equation complexity.

In sum, the comparison between a uniform and a biased universal grammar demonstrates that varying the production probabilities of a probabilistic grammar can reduce the expected number of samples by more than one order of magnitude. Furthermore, the effect is greater for more complex equations. Probabilistic context-free grammars thus provide a powerful and flexible way of expressing prior beliefs and domain knowledge for equation discovery. In the following section, we provide empirical support for this claim.

\begin{table}[t!]
	   \centering
       \caption{Spearman correlation between various measures of equation complexity and speedup, i.e., reduction in the expected number of samples for PCFGs vs. CFGs and for uniform vs. biased PCFGs.}
	   \begin{tabular}{@{}lcr@{}}
	        \\
	        \toprule
	       	\textbf{complexity measure}  & \textbf{prob. vs det.} & \textbf{biased vs uniform}   \\ \midrule
    	    \multicolumn{1}{l}{\textbf{string length}} & 0.79 & 0.50   \\
	       	\multicolumn{1}{l}{\textbf{number of unique variables}} & 0.55 & 0.69 \\
	       	\multicolumn{1}{l}{\textbf{number of operators}} & 0.78 & 0.32 \\ \bottomrule
	       \end{tabular}
        \label{tab:complexity}
\end{table}

\section{Empirical analysis}
To accompany the preceding theoretical analysis, we have developed a simple algorithm for performing equation discovery with probabilistic context-free grammars in Python. We take a Monte-Carlo approach to sampling the probability distribution over the space of all possible equations. An equation discovery task includes simulated or measured data, as well as the names of the variables for which data are provided. In our study of performance, each task is accompanied by the correct equation, to which we can compare the results of the algorithm (the discovered equations).

For each equation discovery task, a universal arithmetic PCFG is constructed with terminal symbols corresponding to the measured variables, as well as the constant symbol. A large number of candidate equation structures are randomly sampled from the grammar and evaluated against the data in the Feynman database. This involves the selection of values of the constants equation to obtain the best fit. The equation fitting the data best is returned as the result. Alternatively, equation structures could be sampled until a satisfactory candidate was found, where a satisfactory degree of fit would be specified in advance.

\subsection{Monte-Carlo sampling algorithm}
Algorithm~\ref{alg:sample_PCFG} presents an outline of the Monte-Carlo approach to grammar-guided equation discovery. Candidate expressions on the right-hand side of the equations are sampled independently by the procedure \textsc{generate\_sample} from Algorithm~\ref{alg:generate_sample} (line~4). We implemented the algorithm in Python using the support for working with grammars provided by the Python Natural Language Toolkit, NLTK~\cite{nltk}. Each sampled expression $e$ is then processed using the Python library for symbolic mathematics, SymPy~\cite{sympy} to obtain its canonical form $e_c$ and identify the list of constant parameters it includes (line~5). We then estimate the values of these constant parameters by minimizing the relative root mean squared error of the equation $v=e_c$ on $D$, as defined in Equation~\ref{eq:rermse} (line~6). To this end, we employ differential evolution (DE)~\cite{DE} as an efficient and proven method of global optimization, with DE parameters set to values similar to those reported in~\cite{luksic2019}. Following the formula from Equation~\ref{eq:rermse}, the algorithm computes the final score of the candidate equation (line~7). Finally, the algorithm reports the list of the sampled equations sorted by increasing error (lines~8 and 9).

\begin{algorithm}[t]
\caption{Monte-Carlo algorithm for grammar-based equation discovery.}
\begin{algorithmic}[1]
  \Require{Probabilistic grammar $G=(\mathcal{N},\mathcal{T},\mathcal{R},S)$ generating arithmetic expressions, number of samples $N$, data set $D$, target variable $v$}
  \Ensure{List of equations $\textit{eqns}$, sorted according to increasing error on $D$}
  \Procedure{MC\_GBED}{$G$, $N$, $D$, $e$}
  \State $\textit{eqns}$ = [\ ]
  \For{$i=1$, $i \leq N$}
  \State $(e, p)$ = \textsc{generate\_sample}($G$, $S$)
  \State $e_c$ = canonical\_form($e$)
  \State $\textit{eqn}$ = fit\_parameters($e_c$, $v$, $D$)
  \State $\textit{error}$ = $\textit{ReRMSE}(\textit{eqn}, D)$
  \State $\textit{eqns}$.append($\textit{eqn}$, $p$, $\textit{error}$)
  \EndFor
  \State \textbf{return} $\textit{eqns}$.sort(key=$\textit{error}$, order=increasing)
  \EndProcedure
\end{algorithmic}
\label{alg:sample_PCFG}
\end{algorithm}


\subsection{Empirical setup}
In order to confirm our theoretical analysis empirically, we employ the described Monte-Carlo algorithm for equation discovery on the Feynman database of one hundred equations from physics. We compare the uniform and the biased versions of the universal arithmetic grammar from Equation~\ref{eq:grammar-universal-pcfg}, with production probability distributions parametrized as in Table~\ref{tab:parameters}. For each Feynman equation, Algorithm~\ref{alg:sample_PCFG} generates $10^5$ candidate expressions with each of the studied grammars. Due to practical concerns and time constraints, parameter estimation is performed only for equations with at most five constant parameters. Equations with more than five constant parameters are simply assumed to be inadmissible.

We further optimize the computational complexity of the equation discovery process by checking the generated expressions for duplicates, i.e., expressions with identical canonical forms; the parameters for all the expressions sharing the same canonical form are estimated only once. Moreover, since the data in the Feynman database are noise free, correct equations tend to have low error. In the experiments, we use $10^{-9}$ as a threshold on the value of ReRMSE to decide whether a candidate is a correct reconstruction of the target equation in the Feynman database: a target equation from the Feynman database is successfully reconstructed with a given grammar, if the corresponding sample contains at least one matching candidate equation. We perform three independent runs of the Monte-Carlo sampling for each grammar to account for the algorithm's stochastic nature.


\subsection{Results}
\begin{table}
	   \centering
       \caption{Summary of experimental results on reconstructing the hundred target equations from the Feynman database using the Monte-Carlo algorithm for grammar-based equation discovery with the uniform and biased versions of the universal grammar for arithmetic expressions.}
	   \begin{tabular}{@{}lccr@{}}
	        \\
	        \toprule
	       	\textbf{Experiment}  & \textbf{N unique [$\cdot 10^3$]} & \textbf{Coverage} & \textbf{Reconstructed equations}  \\ \midrule
    	    \multicolumn{1}{l}{\textbf{uniform 1}} & $31 \pm 4 $ & $0.36 \pm 0.04$ & 36  \\
	       	\multicolumn{1}{l}{\textbf{uniform 2}} & $31 \pm 4$  & $0.36 \pm 0.04$ & 38 \\
	       	\multicolumn{1}{l}{\textbf{uniform 3}} & $28 \pm 7$  & $0.32 \pm 0.08$ & 35 \\
	       	\midrule
	       	\multicolumn{1}{l}{\textbf{biased 1}} &  $31 \pm 4$  & $0.51 \pm 0.04$  & 37 \\ 
	       	\multicolumn{1}{l}{\textbf{biased 2}} &  $30 \pm 5$  & $0.49 \pm 0.07$ & 37 \\
	       	\multicolumn{1}{l}{\textbf{biased 3}} &  $28 \pm 7$  & $0.46 \pm 0.10$ & 36 \\ \bottomrule
	       \end{tabular} 
        \label{tab:results}
\end{table}
Table~\ref{tab:results} summarizes the empirical results. Detailed results for each Feynman equation are reported in the Appendix. The first column in the table reports the number of unique expressions sampled per target equation, averaged over all the problems from the Feynman database. Because of the semantic ambiguity of the grammars for arithmetic expressions, the reported numbers are much lower than the total number of sampled expressions: only about $30\%$ of the samples correspond to unique canonical expressions for both the uniform and the biased grammar. 

The second column in the table reports the achieved coverage of the space of all candidate equations in terms of the total probability of the sampled expressions (i.e., the sum of the probabilities of the corresponding parse trees). Note that sampling with the biased grammar covers around half of the space of candidate equations (in terms of total probability), while the uniform universal grammar reaches a coverage of only around $0.35$. We attribute this difference to changes in the structure of the search space, brought by the different probability distributions of production rules. The probability distributions for the biased grammar are generally more varied --- some rules have higher probabilities, others have lower probabilities, as opposed to equal probabilities for all. The effect is similar when considering the probability distributions over parse trees. For the biased grammar, a few parse trees contribute most of the covered total probability, while the majority of parse trees only make a minuscule contribution. Furthermore, parse trees with a higher contribution to coverage are also more likely to be sampled. In this way, it is possible to interpret coverage as a measure of inequality among the parse trees, which is related to the amount of information in the prior distribution. These observations correspond to the idea that uniform priors are the least informative ones, a concept often used in Bayesian statistics \cite{BayesianStatistics}.  

Finally, the third column reports the number of successfully reconstructed target equations from the Feynman database. The two grammars reconstruct roughly 36 of the 100 Feynman equations after $10^5$ samples. When measured in terms of successful reconstructions, the performance of both grammars is seemingly identical. This is in contrast with our theoretical analysis in Section~3, which suggests that the biased grammar is superior.

We can compare the results in Table~\ref{tab:results} to those reported by Udrescu \cite{tegmark2}. AI Feynman was able to discover all one hundred equations from the Feynman database, while the symbolic regression method Eureqa~\cite{schmidt2009} was able to reconstruct 71\% of the equations. In comparison, probabilistic grammar-based equation discovery reconstructed 37\% of the equations in our experiments. The lower performance is not unexpected, as the method presented in this paper serves mainly an illustrative purpose and relies on a very simple algorithm. We are going to return to this comparison in the analysis of the results.

\subsection{Connection to the theoretical analysis}
The results in Table~\ref{tab:results} can not be immediately matched with the theoretical results, as they summarize performance at only a fixed number ($10^5$) of sampled expressions. To analyze the impact of the number of samples on the performance, we re-sample the expressions in each experiment, as detailed in the Appendix. Figure~\ref{im:successrate} depicts the results of the re-sampling: the y-axis shows the percentage of successfully reconstructed Feynman equations, which increases as the number of sampled equations, shown on the x-axis, increases. The shaded areas depict the range between the minimum and the maximum average success rate, obtained with the three runs of the Monte-Carlo algorithm. These results reveal the difference in performance between the uniform and the biased universal grammar. The performance of the biased grammar increases faster and remains well above the performance of the uniform grammar for sample sizes of up to 23,000 expressions.

\begin{figure}[th]
    \centering
    \includegraphics[width=1\textwidth]{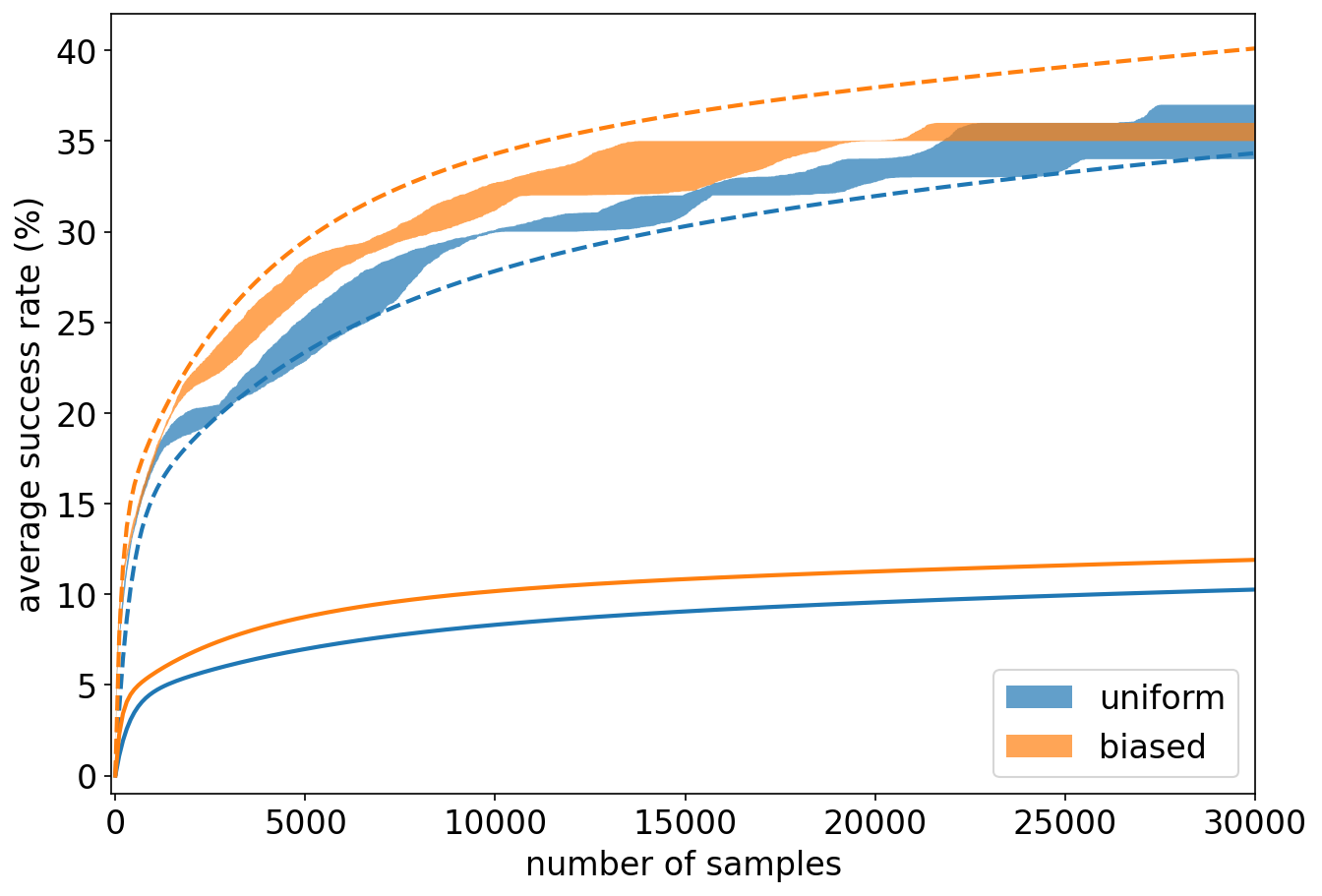}
    \caption{Average rate of successful reconstruction of the Feynman equations achieved with the uniform and biased grammar. The filled region represents empirical results across the three independent runs of Algorithm~2. The solid lines correspond to the predicted success rates based on the analysis in Section~3.4. The dashed lines represent the predicted success rates, corrected by taking into account the empirically estimated level of semantic ambiguity for each grammar.
    \label{im:successrate}}
\end{figure}

The two full lines on the same figure represent the success rates as predicted by the theoretical analysis in Section~3: note that they are much lower than the empirically obtained success rates. The reason for this discrepancy is the semantic ambiguity of the universal grammar. Although we know that the grammars generate many expressions that are mathematically identical, we approximate the sum of their probabilities with the probability of only a single parse tree. We infer the ratio between these two values from our experiments as the ratio between the number of unique expressions (second column in Table~\ref{tab:results}) in the sample and the sample size ($10^5$), averaged over the three independent samplings:
\begin{itemize}
    \item uniform:\quad $\frac{N_{\text{unique}}}{N} = 0.299$,
    \item biased:\quad $\frac{N_{\text{unique}}}{N} = 0.297.$
\end{itemize}
We use these ratios to correct the estimated probability of generating a parse tree that simplifies to a canonical expression corresponding to the solution of the studied problem. This correction reduces the theoretical expectation on the number of necessary samples for each grammar, which leads to the dashed lines in Figure~\ref{im:successrate}. The corrected theoretical predictions of the expected number of samples and the respective success rates for each grammar correspond well to the empirical curves, calculated by re-sampling the empirical results. Note that the corrected predictions provide solid confirmation of our theoretical analysis.

\subsection{Analysis of the results}
\begin{figure}[t!]
    \centering
    \includegraphics[width=1\textwidth]{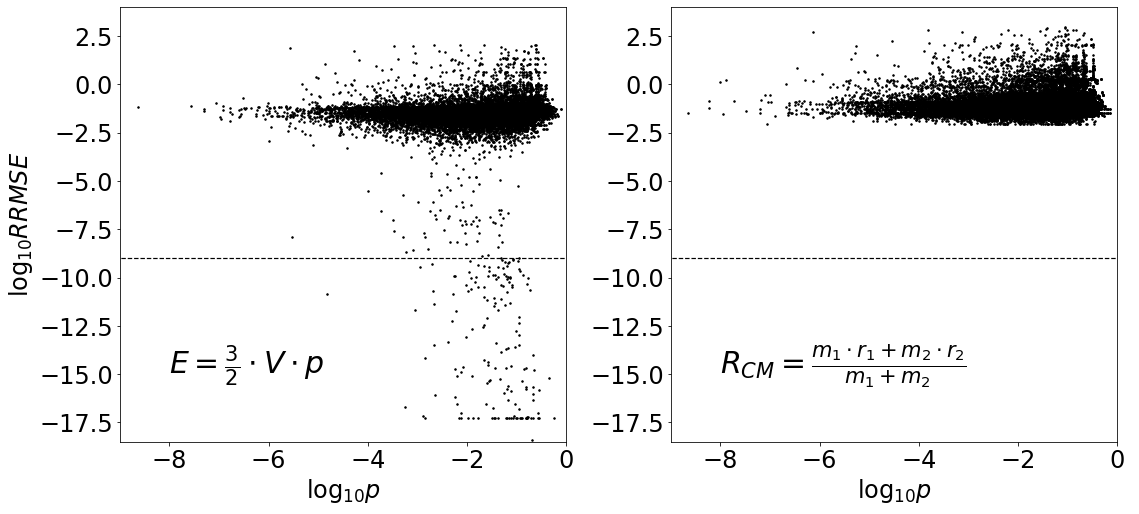}
    \caption{Scatter plots of the probability of a sampled expression against the error of the corresponding equation for two samples taken with the uniform universal grammar. The samples on the left correspond to a simple, successfully reconstructed target equation from the Feynman database, while the samples on the right correspond to a more complex equation that was not successfully reconstructed. The dashed line represents our error threshold for considering a candidate expression to be correct. The best sampled expressions are found in the bottom right corner of each scatter plot---they have high probability and the corresponding equations have low error.
    \label{im:eq-examples}}
\end{figure}
Overall, the performance of the Monte Carlo algorithm for grammar-based equation discovery is not great. In our experiments, the method was able to solve around 37\% of the equation discovery tasks from the Feynman dataset. To understand the achieved performance level, we have to take a closer look at the results for some of the tasks.

In Figure~\ref{im:eq-examples}, we study one of successfully solved equation discovery tasks and one of the unsuccessful ones. We visualize the distributions of the probability of an expression and its error as a scatter plot on a logarithmic scale. We see that, for both equation discovery tasks, the majority of sampled expressions are found in a cluster with moderate probability and high error. For the solved problem, a number of points are scattered across more than 15 orders of magnitude in error, with many of them falling below the error threshold for a correct expression. In the case of the unsolved problem, there are no points at all below the main cluster. It is clear that the solved problem represents an easier task than the unsolved one if we take a look at the corresponding information in table in Appendix~A.7 (equations 20 and 39). Equation \#39 is less complex than \#20 in all measures except the number of parameters. With an estimated generation probability of $3 \cdot 10^{-4},$ at least a few correct solutions for task \#39 are expected in a sample of $10^5$ expressions, on average. In contrast, the estimated probability of $9 \cdot 10^{-15}$ of equation \#20 means that we would need to be extremely lucky to stumble upon the correct solution in our samplings.
\begin{figure}[th]
    \centering
    \includegraphics[width=1\textwidth]{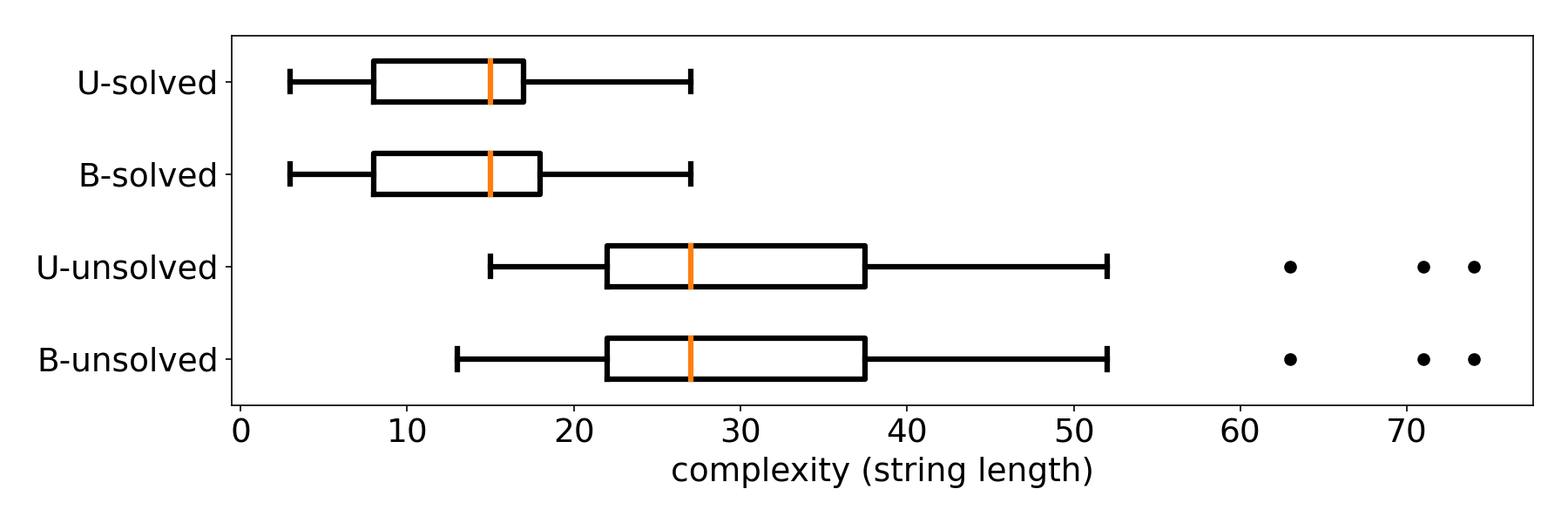}
    \caption{A box plot comparison of the complexity of equations from the Feynman database that were successfully reconstructed in the experiments (solved) with the complexity of equations that the algorithm was unable to reconstruct (unsolved). Depicted separately are experiments using the uniform universal grammar (labelled $U$) and the biased universal grammar (labelled $B$). The orange line indicates the median of the distribution, while dots indicate outliers.
    \label{im:complexity-boxplot}}
\end{figure}

Another high-level view of the results is presented in Figure~\ref{im:complexity-boxplot}, where we compare the distributions of target expression complexity for the sets of solved and unsolved problems from the Feynman database. We see that the majority of problems the Monte-Carlo algorithm was unable to solve are more complex than the majority of solved problems, although there is some overlap in the tails of the distributions. 

The Monte-Carlo algorithm presented in this work is intended to serve as a demonstration of the concepts and ideas behind using probabilistic grammars as a method of generating expressions for equation discovery. Furthermore, it provides a baseline for the performance that can be achieved with arguably the simplest sampling procedure possible. Our complexity analysis reveals where simple Monte-Carlo sampling fails, hinting towards a path for more efficient approaches. The nature of sampling a probabilistic grammar, as well as our chosen prior distributions, focuses the search toward simpler expressions. This bias allows the procedure to successfully discover the solutions to the majority of less complex problems from the Feynman database. During the sampling procedure, however, we ignore all information on the errors of the previously generated expressions. In order for a method, based on sampling  probabilistic grammars, to be able to discover complex equations, intelligent and adaptive data-driven sampling algorithms will have to be developed.

Note finally, that the complexity of a parse tree deriving a particular arithmetic expression is not directly related to the complexity of that expression, as observed in Figure~\ref{im:complexity-boxplot}. The complexity of the parse tree depends on the grammar used for equation discovery: for one grammar, the expression can be derived using a parse tree of depth three, while for another grammar, a parse tree of depth ten might be needed. In this paper, we have used only the universal grammar for general arithmetic expressions from Equation~\ref{eq:grammar-universal-pcfg}. Using alternative probabilistic grammars (for example, derived from the deterministic grammars considered by Todorovski and Džeroski~\cite{lagramge} that employ domain-specific or cross-domain knowledge about mathematical modeling to constrain the space of candidate equations), would allow for much more efficient reconstruction of the target equations from the Feynman database. These alternative grammars might need simpler parse trees to derive the target equations, which would decrease the number of samples required to reconstruct the target equation.

\section{Discussion}
The theoretical and empirical analysis show that the formalism of probabilistic grammars provides a flexible and powerful framework for specifying the inductive bias for equation discovery and symbolic regression. In particular, probabilistic grammars allow for concise specification of the prior distribution over the space of candidate equations. This is achieved by the probability distributions over the production rules for each non-terminal symbol in the grammar. For example, this provides the user of the equation discovery algorithm with an immediate and transparent control over the parsimony principle by specifying the probabilities of the recursive production rules that allow the grammar to generate complex arithmetic expressions. As shown in Section~3, lowering the probability of the recursion rules increases the probability of simpler equations.

This is an important result as the parsimony principle, following the Occam's razor that favors simpler explanations over complex ones, is of crucial importance in algorithms for equation discovery and symbolic regression~\cite{tegmark2019}. It has been so far addressed by using various techniques, including the minimum-description-length formalism~\cite{tegmark2019,peckov2008}, Akaike and Bayesian information criteria~\cite{mangan2017}, regularization methods for sparse regression~\cite{kutz} and complexity-related penalty terms in the fitness for evolutionary approaches~\cite{schmidt2009}. These techniques for following the parsimony principle often require an appropriate setting of a (regularization) parameter that sets the degree of trade-off between the error of the equation and its complexity. Finding the optimal parameter setting might require computationally expensive trial-and-error experiments. In contrast, the frame presented in Section~3 provides a basis for setting the probabilities of the recursive rules analytically, based on the probabilities of simpler expressions corresponding to shallower parse trees.

Grammars can support different types of inductive bias. In the experiment with the biased grammar, we have shown that the probabilities of production rules involving different arithmetic operators can be also finely tuned, which is equivalent to (if not more general than) setting the probabilities of the arithmetic operators~\cite{guimera2020}. Grammars can not only specify \textsl{soft} constraints that increase/decrease the probability of (classes of) expressions, but can also be used to specify \textsl{hard} constraints on the space of candidate equations. As shown by Todorovski and Džeroski~\cite{lagramge}, this kind of hard constraints allows for precise tailoring of the space of candidate equations to the knowledge and basic principles in the domain of interest. Other aspects of cross-domain knowledge~\cite{pret}, including dimensional analysis~\cite{sds}, can also be encoded as grammars~\cite{sebag}.

Incorporating background domain-specific knowledge in equation discovery is closely related to the communicability of the obtained equations with scientists in the domain of interest. Mathematical models and equations, that are in line with the knowledge in the domain of use, allow humans to interpret them in terms of explanations of the observed phenomena~\cite{ipm}. However, deterministic grammars, as well as other types of constraints~\cite{scipm}, as vehicles for integrating knowledge, cut out entire regions of the space of candidate equations. While these cuts improve the computational efficiency, they introduce the risk of cutting off potentially valid models. Probabilistic grammars allow for soft constraints that can reduce the risk by specifying the probability distribution over the space of candidate equations. At the same time, probabilistic grammars can still improve the computational complexity of the equation discovery process and lead to communicable models. The grammar parse trees and their inner nodes correspond to non-terminals and potentially explanatory higher-order expressions, much like processes in explanatory process-based models~\cite{pbm,ipm}.



The empirical results from Section~4 show that the Monte-Carlo algorithm for grammar-based equation discovery is limited to the reconstruction of simple equations and fails to reconstruct a major fraction of more complex equations from the Feynman database. This result is expected and can be explained with the simplicity of the proposed algorithm, where more elaborate algorithms are expected to lead to better results~\cite{tegmark}. Approaches of Bayesian optimization~\cite{jones1998} can be applied to guide the simple sampling algorithm towards the candidate equations with small error and therefore towards the target equation in the case of reconstruction. We expect that these algorithms will improve the efficiency of the simple sampling procedure presented here and make the sampling-based approaches competitive with current probabilistic and Bayesian approaches to symbolic regression~\cite{guimera2020}. In addition, Bayesian optimization allows for explicit control over the trade-off between exploitation and exploration in the search space of candidate equations~\cite{jones1998}, which has a significant impact on the efficiency of search algorithms for equation discovery~\cite{tanevski2020}.

Probabilistic grammars can be also employed in the context of learning bias for equation discovery and process-based modeling~\cite{bridewell2010}. In that setting, constraints on the space of candidate equations are inferred from the results of an equation discovery algorithm: we learn constraints that discriminate between accurate and inaccurate equations. Similarly, the Bayesian scientist~\cite{guimera2020} learns or estimates the probabilities of different arithmetic operators from a corpus of equations from Wikipedia. In both cases, the induced bias for equation discovery can be then transferred from one discovery task or domain to another. The transfer can significantly improve the efficiency of symbolic regression without harming the accuracy of the discovered equations. However, the formalism of process-based modeling requires complex learning mechanisms in logic~\cite{bridewell2010}. When using probabilistic context free grammars as inductive bias, the latter can be learned using a number of standard algorithms for inferring grammars~\cite{higuera2005}.



\section{Conclusion}
The results presented in this paper confirm the validity of our conjecture that the use of probabilistic grammars has a remarkable impact on the computational efficiency of methods for equation discovery and symbolic regression. First, our theoretical analysis shows that probabilistic grammars can significantly reduce the expected number of equations needed to reconstruct a known equation from data. Second, probabilistic grammars provide an intuitive, probabilistic parametrization of the parsimony principle, that favors simpler equations over more complex ones, through the probabilities of the recursive production rules in the grammar. Third, the computational experiments with a simple Monte-Carlo sampling algorithm for equation discovery provide an empirical confirmation of the theoretical results.

The significance of the contributions presented in this paper is related to the opportunities for further development of algorithms for equation discovery and symbolic regression. The integration of probabilistic grammars into approaches to equation discovery enables the development of truly Bayesian approaches to equation discovery. In particular, this development will focus on advanced and more efficient sampling procedures that will guide the search for proper equations towards more promising parts of the space of candidate equations specified by the grammar.  The computational efficiency of equation discovery algorithms can be further improved by encoding background, domain-specific knowledge into probabilistic grammars. This encoding can be pursued with manual labor of transforming domain knowledge into grammars or by employing computational approaches to grammar inference from examples of equations in the particular domain of interest.

\section*{Acknowledgements}
The authors acknowledge the financial support of the Slovenian Research Agency via the research core funding No.~P2-0103, No.~P5-0093 and projects No.~N2-0128, No.~V5-1930. We would also like to thank Jovan Tanevski for providing feedback on early drafts of the manuscript.

\bibliographystyle{acm}
\bibliography{references.bib}

\appendix
\section{Appendix}
\subsection{Analytical expression for the number of parse trees with a given height for grammar $G_L$}
The grammar in Equation \ref{eq:grammar_linear} is simple enough to derive the number of parse trees with height $h$ analytically, using Equation \ref{eq:counting-n}. First, observe that there are only two trees with $V$ as the root and their height is one:
\begin{equation}
    n_{G_L}(V, h) = 
    \begin{cases}
    n_V & \text{if $h=1$}\\
    0 & \text{otherwise.}
\end{cases}
\end{equation}
Consequently, $N_{G_L}(V, h) = n_V$ for all $h \ge 1$. We now make use of this while considering Equation \ref{eq:counting-n} for $h>=2$ and $E$ as the root node:
\begin{align}
    n_{G_L}(E, h) &= n_V - n_V +  n_V \left( N_{G_L}(E, h-1) - N_{G_L}(E, h-2) \right) \\
    n_{G_L}(E, h) &= n_V \left( N_{G_L}(E, h-1) - N_{G_L}(E, h-2) \right) \\
    n_{G_L}(E, h) &= n_V \times n_{G_L}(E, h-1).
\end{align}
This relation is simply the recursive form of a geometric sequence. The general form can be written as 
\begin{equation}
    n_{G_L}(E, h) = 
    \begin{cases}
    n_V^{h-1} & \text{if $h \ge 2$} \\
    0 & \text{otherwise.}
    \end{cases}
\end{equation}
The number of parse trees with height up to and including $h$ is then the sum of the first $h$ terms of the geometric series:
\begin{equation}
    N_{G_L}(E, h) = \sum\limits_{h_i=2}^h n_{G_L}(E, h_i) = \sum\limits_{h_i=2}^h n_V^{h_i} = \frac{n_v^h-1}{n_v-1}-1.
\end{equation}
In our computations we use a $G_L$ with $n_V = 2$, for which the expression simplifies to $N_{G_L}(E, h) = 2^h - 2$.

\subsection{Analytical expression for coverage for grammar $G_L$}
The probability of generating a parse tree with height $h$ with grammar $G_L$ is 
$$P_{G_L}(E, h) = p^{h-2}(1-p).$$
This is easy to see by considering the Bernoulli process. Recall that one of the productions with $E$ on the left-hand side represents recursion, with probability $p$, and the other ends the process of generation, with probability $1-p$. Starting from 2, the generator must choose the recursive option $h-2$ times before choosing the nonrecursive production in order to generate a parse tree with height $h$. Then, we can compute the coverage as
$$Cov_{G_L}(E, h) = \sum\limits_{h_i=2}^h p^{h_i-2}(1-p) = (1-p)\sum\limits_{h_i = 0}^{h-2}p^{h_i} = \frac{1-p^{h-1}}{1-p}(1-p) = 1 - p^{h-1}.$$
To demonstrate the usage of the recursive Equation \ref{eq:counting-Cov}, we can use it to compute the above result. First, we observe the coverage for parse trees with $V$ as their root node:
$$Cov_{G_L}(V, h) = 1,\ \text{if $h \ge 1$}.$$
From there, we can derive a recursive relation for the coverage of parse trees with $E$ as the root node:
\begin{align}
Cov_{G_L}(E, h) &= p \cdot Cov_{G_L}(E, h-1)\cdot Cov_{G_L}(V, h-1) \\
\phantom{Cov_{G_L}(E, h)} &\phantom{=} + (1-p)\cdot Cov_{G_L}(V, h-1) \\
Cov_{G_L}(E, h) &= p \cdot Cov_{G_L}(E, h-1) + (1-p).
\end{align}
Here it is not as easy to see the general expression for the $h$-th term in the series. Instead, we consider the first three terms:
\begin{align}
    Cov_{G_L}(E, 1) &= 0, \\
    Cov_{G_L}(E, 2) &= 1-p, \\
    Cov_{G_L}(E, 3) &= p(1-p) +(1-p) = 1-p^2.
\end{align}
From this sequence it is easy to guess the general form $Cov_{G_L}(E, h) = 1 - p^{h-1}.$ For proof by induction, we now assume the relation holds for $h$ and use the recursive relation to prove it for $h+1$:
$$Cov_{G_L}(E, h+1) = p\cdot Cov_{G_L}(E, h) + 1-p =  p (1 - p^{h-1}) + 1-p  = 1 - p^h.$$

%

\subsection{Resampling procedure}
In the following Appendix section, we elaborate on the details of the post processing we employed in order to create Figure~\ref{im:successrate}.
In the experiment, described in Section~4, we employ the developed Monte-Carlo algorithm for grammar-based equation discovery to generate a large number of candidate expressions using the universal grammar with either a uniform or a biased prior distribution. Using each grammar we perform three independent samplings of $10^5$ expressions with unique parse trees. Due to semantic ambiguity, some of these expressions simplify to the same canonical expression. We end up with approximately 30000 unique canonical expressions for each of the equation discovery tasks from the Feynman database. 

Analyzing the sampled expressions and drawing meaningful conclusions is not a simple task. In our analysis, we characterize each sampled expression by two values: its generation probability and the relative root mean squared error (RRMSE) it produces when evaluated with the optimal parameter values. We set RRMSE=$10^{-9}$ as the threshold for a correct solution. We consider an equation discovery task successful if the Monte-Carlo sample of expressions contains at least one expression with error under the threshold. When agglomerating the results of three independent samplings, the number of successes for a particular problem can range from zero to three successes. In Table~A.8 we report this value in columns \#$S_U$ and \#$S_B$.

The results obtained in this way only serve to inform us of the performance of the algorithm at a particular number of samples, $10^5$ in our case. A more general presentation of the results would show a performance curve, in other words, performance vs. sample size. The easiest way would be to simply take a sliding minimum across the error of the sampled expressions in the order they were sampled. However, the order is arbitrary - if we repeat the experiment, we would get the sampled expressions in an entirely different order (based on their probabilities) and a different performance curve. To alleviate the stochasticity of the algorithm, we employ resampling on the sample of expressions. The measure of success we are working with is the success ratio,
which we define as the portion of successfully reconstructed equations from the Feynman dataset. 
We treat the sample of approx. 30000 unique canonical expressions with their corresponding probabilities as a probability distribution. From this distribution we then sample 30000 expressions, without replacement. In other words, we randomly reorder the set of sampled expressions, while taking into account the probability of each expressions. By repeating this many times (100 in our case), we can average the success ratio at each sample size. 
This value is the average success rate, depicted in Figure~\ref{im:successrate}. It can be interpreted as the expected success rate at a given sample size. In other words, if the Monte-Carlo algorithm were repeatedly run many times, each time sampling N canonical expressions for each problem from the Feynman database, computing the average of the portion of solved problems would give a value close to the average success rate we report.

In our experiment, we perform three independent Monte-Carlo samplings for each of the two grammars, resulting in six sets of sampled expressions. The resampling procedure is performed separately for each of the six sample sets. Figure~\ref{im:successrate} depicts the minimum and maximum of the average success rate across the three samplings at each sample size for each of the two grammars.

\subsection{Theoretical expectation of success rate}
In the theoretical analysis in Section~3, we use an inside chart parser algorithm \cite{manning1999} to parse the target expression of each problem from the Feynman database, using either the universal or the biased universal grammar. We take the parsed tree probabilities as an approximation for the probability that a randomly sampled expression corresponds to the correct solution for a given problem. We can use the probabilities to calculate a theoretical expectation of the success rate (dependent on the sample size) for each grammar. The probability of finding the correct solution of a problem with index $i$ in a sample of $N$ expression, generated with grammar G, is
$$ P_{G,i}(N) = 1 - (1-p_{G,i})^N \approx  1 - (1 - \widetilde{p_{G,i}})^N,$$
where $p_{G,i}$ is the probability of randomly generating the correct solution to problem $i$ using grammar G. We approximate this probability with the parsed probability of a single parse tree $p_{G,i} \approx \widetilde{p_{G,i}}$. 
Finally we consider the complete Feynman dataset of one hundred equations to arrive at the expected success rate
$$ E[\text{success rate}](N) = \frac{1}{100}\sum\limits_{i=1}^{100}P_{G,i}(N)
\approx 1-\frac{1}{100}\sum\limits_{i=1}^{100}(1-\widetilde{p_{G,i}})^N.$$
Once we perform our sampling experiment, we can directly compare the empirical success rate with its theoretical expectation, which is depicted as the pair of full lines in Figure~\ref{im:successrate}.
\subsection{Feynman database}
The Feynman database was constructed by Udrescu and Tegmark \cite{tegmark} to facilitate the development and testing of algorithms for symbolic regression. The database is composed of one hundred important equations from physics and acts as a good playground and benchmark dataset for equation discovery. The following table presents the one hundred equations from the Feynman database, along with their file name, corresponding to the presentation in \cite{tegmark}.

\begin{landscape}
\begin{longtable}{lll}
\toprule
{} &   Filename &                                                                     Formula \\
\midrule
\endhead
\midrule
\multicolumn{3}{r}{{Continued on next page}} \\
\midrule
\endfoot

\bottomrule
\endlastfoot
0  &     I.6.2a &                                                 exp(-theta**2/2)/sqrt(2*pi) \\
1  &      I.6.2 &                                 exp(-(theta/sigma)**2/2)/(sqrt(2*pi)*sigma) \\
2  &     I.6.2b &                        exp(-((theta-theta1)/sigma)**2/2)/(sqrt(2*pi)*sigma) \\
3  &     I.8.14 &                                                 sqrt((x2-x1)**2+(y2-y1)**2) \\
4  &     I.9.18 &                                  G*m1*m2/((x2-x1)**2+(y2-y1)**2+(z2-z1)**2) \\
5  &     I.10.7 &                                                       m\_0/sqrt(1-v**2/c**2) \\
6  &    I.11.19 &                                                           x1*y1+x2*y2+x3*y3 \\
7  &     I.12.1 &                                                                       mu*Nn \\
8  &     I.12.2 &                                                 q1*q2*r/(4*pi*epsilon*r**3) \\
9  &     I.12.4 &                                                    q1*r/(4*pi*epsilon*r**3) \\
10 &     I.12.5 &                                                                       q2*Ef \\
11 &    I.12.11 &                                                       q*(Ef+B*v*sin(theta)) \\
12 &     I.13.4 &                                                      1/2*m*(v**2+u**2+w**2) \\
13 &    I.13.12 &                                                         G*m1*m2*(1/r2-1/r1) \\
14 &     I.14.3 &                                                                       m*g*z \\
15 &     I.14.4 &                                                           1/2*k\_spring*x**2 \\
16 &    I.15.3x &                                                   (x-u*t)/sqrt(1-u**2/c**2) \\
17 &    I.15.3t &                                              (t-u*x/c**2)/sqrt(1-u**2/c**2) \\
18 &     I.15.1 &                                                     m\_0*v/sqrt(1-v**2/c**2) \\
19 &     I.16.6 &                                                          (u+v)/(1+u*v/c**2) \\
20 &     I.18.4 &                                                       (m1*r1+m2*r2)/(m1+m2) \\
21 &    I.18.12 &                                                              r*F*sin(theta) \\
22 &    I.18.14 &                                                            m*r*v*sin(theta) \\
23 &     I.24.6 &                                        1/2*m*(omega**2+omega\_0**2)*1/2*x**2 \\
24 &    I.25.13 &                                                                         q/C \\
25 &     I.26.2 &                                                       arcsin(n*sin(theta2)) \\
26 &     I.27.6 &                                                               1/(1/d1+n/d2) \\
27 &     I.29.4 &                                                                     omega/c \\
28 &    I.29.16 &                                sqrt(x1**2+x2**2-2*x1*x2*cos(theta1-theta2)) \\
29 &     I.30.3 &                                     Int\_0*sin(n*theta/2)**2/sin(theta/2)**2 \\
30 &     I.30.5 &                                                         arcsin(lambd/(n*d)) \\
31 &     I.32.5 &                                               q**2*a**2/(6*pi*epsilon*c**3) \\
32 &    I.32.17 &     (1/2*epsilon*c*Ef**2)*(8*pi*r**2/3)*(omega**4/(omega**2-omega\_0**2)**2) \\
33 &     I.34.8 &                                                                     q*v*B/p \\
34 &     I.34.1 &                                                             omega\_0/(1-v/c) \\
35 &    I.34.14 &                                           (1+v/c)/sqrt(1-v**2/c**2)*omega\_0 \\
36 &    I.34.27 &                                                            (h/(2*pi))*omega \\
37 &     I.37.4 &                                              I1+I2+2*sqrt(I1*I2)*cos(delta) \\
38 &    I.38.12 &                                         4*pi*epsilon*(h/(2*pi))**2/(m*q**2) \\
39 &     I.39.1 &                                                                    3/2*pr*V \\
40 &    I.39.11 &                                                             1/(gamm-1)*pr*V \\
41 &    I.39.22 &                                                                    n*kb*T/V \\
42 &     I.40.1 &                                                      n\_0*exp(-m*g*x/(kb*T)) \\
43 &    I.41.16 &             h/(2*pi)*omega**3/(pi**2*c**2*(exp((h/(2*pi))*omega/(kb*T))-1)) \\
44 &    I.43.16 &                                                           mu\_drift*q*Volt/d \\
45 &    I.43.31 &                                                                    mob*kb*T \\
46 &    I.43.43 &                                                           1/(gamm-1)*kb*v/A \\
47 &     I.44.4 &                                                            n*kb*T*ln(V2/V1) \\
48 &    I.47.23 &                                                           sqrt(gamm*pr/rho) \\
49 &     I.48.2 &                                                    m*c**2/sqrt(1-v**2/c**2) \\
50 &    I.50.26 &                                     x1*(cos(omega*t)+alpha*cos(omega*t)**2) \\
51 &    II.2.42 &                                                           kappa*(T2-T1)*A/d \\
52 &    II.3.24 &                                                             Pwr/(4*pi*r**2) \\
53 &    II.4.23 &                                                          q/(4*pi*epsilon*r) \\
54 &    II.6.11 &                                        1/(4*pi*epsilon)*p\_d*cos(theta)/r**2 \\
55 &   II.6.15a &                                 p\_d/(4*pi*epsilon)*3*z/r**5*sqrt(x**2+y**2) \\
56 &   II.6.15b &                             p\_d/(4*pi*epsilon)*3*cos(theta)*sin(theta)/r**3 \\
57 &     II.8.7 &                                                   3/5*q**2/(4*pi*epsilon*d) \\
58 &    II.8.31 &                                                             epsilon*Ef**2/2 \\
59 &    II.10.9 &                                                 sigma\_den/epsilon*1/(1+chi) \\
60 &    II.11.3 &                                              q*Ef/(m*(omega\_0**2-omega**2)) \\
61 &   II.11.17 &                                            n\_0*(1+p\_d*Ef*cos(theta)/(kb*T)) \\
62 &   II.11.20 &                                                    n\_rho*p\_d**2*Ef/(3*kb*T) \\
63 &   II.11.27 &                                          n*alpha/(1-(n*alpha/3))*epsilon*Ef \\
64 &   II.11.28 &                                                   1+n*alpha/(1-(n*alpha/3)) \\
65 &   II.13.17 &                                                 1/(4*pi*epsilon*c**2)*2*I/r \\
66 &   II.13.23 &                                                   rho\_c\_0/sqrt(1-v**2/c**2) \\
67 &   II.13.34 &                                                 rho\_c\_0*v/sqrt(1-v**2/c**2) \\
68 &    II.15.4 &                                                           -mom*B*cos(theta) \\
69 &    II.15.5 &                                                          -p\_d*Ef*cos(theta) \\
70 &   II.21.32 &                                                  q/(4*pi*epsilon*r*(1-v/c)) \\
71 &   II.24.17 &                                              sqrt(omega**2/c**2-pi**2/d**2) \\
72 &   II.27.16 &                                                             epsilon*c*Ef**2 \\
73 &   II.27.18 &                                                               epsilon*Ef**2 \\
74 &   II.34.2a &                                                                q*v/(2*pi*r) \\
75 &    II.34.2 &                                                                     q*v*r/2 \\
76 &   II.34.11 &                                                                g\_*q*B/(2*m) \\
77 &  II.34.29a &                                                                q*h/(4*pi*m) \\
78 &  II.34.29b &                                                      g\_*mom*B*Jz/(h/(2*pi)) \\
79 &   II.35.18 &                                  n\_0/(exp(mom*B/(kb*T))+exp(-mom*B/(kb*T))) \\
80 &   II.35.21 &                                                n\_rho*mom*tanh(mom*B/(kb*T)) \\
81 &   II.36.38 &                              mom*H/(kb*T)+(mom*alpha)/(epsilon*c**2*kb*T)*M \\
82 &    II.37.1 &                                                               mom*(1+chi)*B \\
83 &    II.38.3 &                                                                     Y*A*x/d \\
84 &   II.38.14 &                                                             Y/(2*(1+sigma)) \\
85 &   III.4.32 &                                          1/(exp((h/(2*pi))*omega/(kb*T))-1) \\
86 &   III.4.33 &                           (h/(2*pi))*omega/(exp((h/(2*pi))*omega/(kb*T))-1) \\
87 &   III.7.38 &                                                          2*mom*B/(h/(2*pi)) \\
88 &   III.8.54 &                                                    sin(E\_n*t/(h/(2*pi)))**2 \\
89 &   III.9.52 &  (p\_d*Ef*t/(h/(2*pi)))*sin((omega-omega\_0)*t/2)**2/((omega-omega\_0)*t/2)**2 \\
90 &  III.10.19 &                                                 mom*sqrt(Bx**2+By**2+Bz**2) \\
91 &  III.12.43 &                                                                n*(h/(2*pi)) \\
92 &  III.13.18 &                                                     2*E\_n*d**2*k/(h/(2*pi)) \\
93 &  III.14.14 &                                                  I\_0*(exp(q*Volt/(kb*T))-1) \\
94 &  III.15.12 &                                                            2*U*(1-cos(k*d)) \\
95 &  III.15.14 &                                                  (h/(2*pi))**2/(2*E\_n*d**2) \\
96 &  III.15.27 &                                                            2*pi*alpha/(n*d) \\
97 &  III.17.37 &                                                    bet*(1+alpha*cos(theta)) \\
98 &  III.19.51 &                        -m*q**4/(2*(4*pi*epsilon)**2*(h/(2*pi))**2)*(1/n**2) \\
99 &  III.21.20 &                                                          -rho\_c\_0*q*A\_vec/m \\
\end{longtable}

\end{landscape}

\subsection{Results}
The table provided in this section presents detailed results of the experiment described in Section~4. Each row corresponds to an equation discovery task  from the Feynman database. For each problem, we performed six independent samplings of $10^5$ candidate equations. Three of these were based on the uniform universal grammar (labelled $U$) and three were based on the biased universal grammar (labelled $B$). The columns of the table, from left to right are as follows;
\begin{itemize}
    \item Index of the problem from the Feynman dataset.
    \item \#v. Number of variables in the target expression.
    \item \#p. Number of constant parameters in the target expression.
    \item \#o. Number of mathematical operations or special function in the target expression.
    \item \#c. Number of characters in the string representation of the target expression.
    \item $p_U$. The probability of generating the target expression, using the uniform universal grammar. Approximated by the probability of a single parse tree. In other words, the approximation ignores the semantic ambiguity of the grammar. 
    \item $p_B$. Same as $p_U$, but using the biased universal grammar.
    \item \#$S_U$. Number of successes in three independent samplings, using the uniform universal grammar. A sampling is successful if it finds at least one model with $RRMSE < 10^{-9}.$
    \item \#$S_B$. Same as \#$S_U$, but using the biased universal grammar.
    \item \#$N_U$. Number of unique expressions in the canonical form, generated using the uniform universal grammar. Formatted as a triplet of values, each corresponding to one of three independent samplings. Expressed in the units of thousands.
    \item \#$N_B$. Same as \#$N_U$, but using the biased universal grammar.
    \item cov$_U$. Sum of probabilities (coverage) of all unique expressions, sampling using the uniform universal grammar. Formatted as a triplet of values, each corresponding to one of three independent samplings.
    \item cov$_B$. Same as cov$_U$, but using the biased universal grammar.
\end{itemize}

\begin{landscape}
\begin{longtable}{lrrrrllrrllll}
\toprule
{} &  \#v &  \#p &  \#o &  \#c &                 $p_U$ &                 $p_B$ &  \#$S_U$ &  \#$S_B$ & $N_U$[$10^3]$ & $N_B$[$10^3]$ &             cov$_U$ &             cov$_B$ \\
\midrule
\endhead
\midrule
\multicolumn{13}{r}{{Continued on next page}} \\
\midrule
\endfoot

\bottomrule
\endlastfoot
0  &    1 &    2 &    6 &   27 &   $1.1{\cdot}10^{-6}$ &   $2.1{\cdot}10^{-7}$ &        3 &        2 &  (30, 30, 30) &   (14, 24, 3) &  (0.38, 0.38, 0.38) &  (0.21, 0.54, 0.29) \\
1  &    2 &    2 &    8 &   43 &  $1.7{\cdot}10^{-11}$ &    $7{\cdot}10^{-12}$ &        0 &        0 &  (35, 35, 25) &  (30, 30, 20) &  (0.38, 0.38, 0.31) &  (0.54, 0.54, 0.34) \\
2  &    3 &    2 &    9 &   52 &  $1.2{\cdot}10^{-19}$ &  $1.5{\cdot}10^{-18}$ &        0 &        0 &  (38, 38, 38) &  (34, 25, 35) &  (0.37, 0.37, 0.37) &  (0.52, 0.47, 0.52) \\
3  &    4 &    0 &    4 &   27 &  $2.2{\cdot}10^{-24}$ &  $7.7{\cdot}10^{-21}$ &        0 &        0 &  (40, 41, 30) &  (38, 37, 28) &   (0.36, 0.36, 0.3) &  (0.51, 0.51, 0.45) \\
4  &    9 &    0 &    8 &   42 &  $6.6{\cdot}10^{-46}$ &  $2.5{\cdot}10^{-39}$ &        0 &        0 &  (46, 46, 46) &  (46, 46, 46) &  (0.33, 0.33, 0.33) &  (0.47, 0.47, 0.47) \\
5  &    3 &    1 &    4 &   21 &  $1.7{\cdot}10^{-11}$ &  $3.8{\cdot}10^{-11}$ &        0 &        0 &  (38, 38, 38) &  (34, 34, 34) &  (0.37, 0.37, 0.37) &  (0.52, 0.52, 0.52) \\
6  &    6 &    0 &    5 &   17 &  $3.6{\cdot}10^{-12}$ &  $3.8{\cdot}10^{-11}$ &        0 &        0 &  (44, 44, 44) &  (42, 42, 42) &  (0.35, 0.35, 0.35) &  (0.49, 0.49, 0.49) \\
7  &    2 &    0 &    1 &    5 &   $2.9{\cdot}10^{-3}$ &   $8.8{\cdot}10^{-3}$ &        3 &        3 &  (35, 35, 35) &  (30, 31, 30) &  (0.37, 0.37, 0.37) &  (0.54, 0.54, 0.53) \\
8  &    4 &    1 &    6 &   27 &  $2.7{\cdot}10^{-15}$ &  $1.7{\cdot}10^{-13}$ &        0 &        1 &  (40, 40, 41) &  (37, 38, 28) &  (0.36, 0.36, 0.36) &  (0.51, 0.51, 0.43) \\
9  &    3 &    1 &    5 &   24 &  $7.5{\cdot}10^{-13}$ &  $2.5{\cdot}10^{-11}$ &        1 &        3 &  (38, 38, 38) &  (34, 35, 35) &  (0.37, 0.37, 0.37) &  (0.52, 0.52, 0.52) \\
10 &    2 &    0 &    1 &    5 &   $2.9{\cdot}10^{-3}$ &   $8.8{\cdot}10^{-3}$ &        3 &        3 &  (35, 35, 30) &  (30, 30, 10) &  (0.38, 0.37, 0.36) &  (0.53, 0.53, 0.07) \\
11 &    5 &    0 &    5 &   21 &  $2.3{\cdot}10^{-13}$ &    $4{\cdot}10^{-12}$ &        0 &        0 &  (42, 42, 30) &  (40, 30, 30) &  (0.35, 0.35, 0.24) &   (0.5, 0.38, 0.39) \\
12 &    4 &    1 &    5 &   22 &  $9.6{\cdot}10^{-16}$ &  $2.3{\cdot}10^{-14}$ &        0 &        0 &  (40, 40, 40) &  (38, 38, 27) &  (0.36, 0.36, 0.36) &  (0.51, 0.51, 0.31) \\
13 &    5 &    2 &    6 &   19 &            $10^{-13}$ &  $6.8{\cdot}10^{-13}$ &        0 &        0 &  (43, 42, 32) &  (40, 40, 40) &  (0.35, 0.35, 0.26) &     (0.5, 0.5, 0.5) \\
14 &    3 &    0 &    2 &    5 &   $3.4{\cdot}10^{-5}$ &     $2{\cdot}10^{-4}$ &        3 &        3 &  (38, 38, 28) &  (34, 25, 34) &  (0.37, 0.37, 0.27) &  (0.52, 0.45, 0.52) \\
15 &    2 &    1 &    3 &   17 &   $1.3{\cdot}10^{-7}$ &   $9.4{\cdot}10^{-7}$ &        3 &        3 &  (35, 35, 25) &  (30, 31, 30) &  (0.37, 0.37, 0.25) &  (0.54, 0.54, 0.53) \\
16 &    4 &    1 &    6 &   25 &  $1.4{\cdot}10^{-17}$ &  $3.3{\cdot}10^{-16}$ &        0 &        0 &  (40, 41, 31) &  (38, 38, 37) &  (0.36, 0.36, 0.23) &  (0.51, 0.51, 0.51) \\
17 &    4 &    1 &    7 &   30 &  $5.5{\cdot}10^{-21}$ &  $2.2{\cdot}10^{-19}$ &        0 &        0 &  (41, 41, 31) &  (38, 27, 18) &   (0.36, 0.36, 0.3) &  (0.51, 0.33, 0.26) \\
18 &    3 &    1 &    5 &   23 &  $4.5{\cdot}10^{-13}$ &  $1.9{\cdot}10^{-12}$ &        0 &        0 &   (38, 38, 8) &  (35, 35, 24) &  (0.37, 0.37, 0.03) &  (0.52, 0.52, 0.37) \\
19 &    3 &    1 &    5 &   18 &  $3.9{\cdot}10^{-15}$ &  $3.6{\cdot}10^{-14}$ &        0 &        0 &  (38, 28, 30) &  (35, 34, 24) &  (0.37, 0.29, 0.26) &  (0.52, 0.52, 0.38) \\
20 &    4 &    0 &    5 &   21 &  $8.6{\cdot}10^{-15}$ &  $3.7{\cdot}10^{-13}$ &        0 &        0 &  (40, 41, 20) &  (37, 37, 10) &  (0.36, 0.36, 0.18) &  (0.51, 0.51, 0.13) \\
21 &    3 &    0 &    3 &   14 &   $4.9{\cdot}10^{-7}$ &   $1.8{\cdot}10^{-6}$ &        3 &        3 &  (38, 38, 28) &  (34, 35, 34) &  (0.37, 0.37, 0.25) &  (0.52, 0.52, 0.52) \\
22 &    4 &    0 &    4 &   16 &   $4.1{\cdot}10^{-9}$ &   $2.9{\cdot}10^{-8}$ &        1 &        2 &  (40, 41, 20) &  (38, 37, 10) &  (0.36, 0.36, 0.22) &  (0.51, 0.51, 0.21) \\
23 &    4 &    1 &    7 &   36 &  $1.6{\cdot}10^{-15}$ &  $7.7{\cdot}10^{-14}$ &        0 &        0 &  (41, 41, 31) &  (38, 37, 28) &  (0.36, 0.36, 0.18) &  (0.51, 0.51, 0.37) \\
24 &    2 &    1 &    1 &    3 &   $2.9{\cdot}10^{-3}$ &   $5.9{\cdot}10^{-3}$ &        3 &        3 &  (35, 25, 30) &  (30, 31, 30) &  (0.38, 0.32, 0.36) &  (0.54, 0.54, 0.54) \\
25 &    2 &    0 &    4 &   21 &                   $0$ &                   $0$ &        0 &        0 &  (35, 35, 35) &  (30, 21, 30) &  (0.38, 0.37, 0.37) &  (0.53, 0.19, 0.53) \\
26 &    3 &    2 &    4 &   13 &   $1.9{\cdot}10^{-9}$ &  $9.5{\cdot}10^{-10}$ &        1 &        0 &  (38, 38, 38) &  (34, 34, 24) &  (0.37, 0.37, 0.37) &  (0.52, 0.52, 0.36) \\
27 &    2 &    0 &    1 &    7 &   $2.9{\cdot}10^{-3}$ &   $5.9{\cdot}10^{-3}$ &        3 &        3 &  (35, 35, 25) &  (30, 30, 30) &  (0.37, 0.37, 0.33) &  (0.53, 0.53, 0.53) \\
28 &    4 &    1 &    8 &   44 &  $1.7{\cdot}10^{-19}$ &  $3.2{\cdot}10^{-18}$ &        0 &        0 &  (40, 31, 40) &  (38, 38, 28) &  (0.36, 0.24, 0.36) &  (0.51, 0.51, 0.39) \\
29 &    3 &    4 &    7 &   39 &    $3{\cdot}10^{-23}$ &  $1.7{\cdot}10^{-24}$ &        0 &        0 &  (38, 38, 28) &  (34, 34, 24) &   (0.37, 0.37, 0.3) &  (0.52, 0.52, 0.46) \\
30 &    3 &    0 &    4 &   19 &                   $0$ &                   $0$ &        0 &        0 &  (38, 38, 20) &  (35, 35, 35) &  (0.37, 0.37, 0.22) &  (0.52, 0.52, 0.52) \\
31 &    4 &    1 &    5 &   29 &  $5.3{\cdot}10^{-17}$ &  $6.5{\cdot}10^{-15}$ &        0 &        0 &  (41, 41, 20) &  (38, 37, 38) &  (0.36, 0.36, 0.22) &  (0.51, 0.51, 0.51) \\
32 &    6 &    1 &   11 &   71 &  $1.9{\cdot}10^{-39}$ &  $3.6{\cdot}10^{-34}$ &        0 &        0 &  (43, 44, 44) &  (42, 42, 32) &  (0.35, 0.35, 0.35) &  (0.49, 0.49, 0.35) \\
33 &    4 &    0 &    3 &    7 &   $2.9{\cdot}10^{-7}$ &   $2.2{\cdot}10^{-6}$ &        3 &        3 &  (41, 40, 21) &  (37, 30, 37) &  (0.36, 0.36, 0.22) &  (0.51, 0.39, 0.51) \\
34 &    3 &    1 &    3 &   15 &   $2.4{\cdot}10^{-8}$ &   $8.7{\cdot}10^{-8}$ &        0 &        0 &  (38, 38, 28) &  (35, 24, 35) &  (0.37, 0.37, 0.15) &  (0.52, 0.46, 0.52) \\
35 &    3 &    2 &    7 &   33 &  $8.2{\cdot}10^{-18}$ &  $2.7{\cdot}10^{-17}$ &        0 &        0 &  (38, 39, 10) &  (34, 34, 35) &  (0.37, 0.37, 0.09) &  (0.52, 0.52, 0.52) \\
36 &    2 &    1 &    3 &   16 &   $2.3{\cdot}10^{-4}$ &   $2.3{\cdot}10^{-4}$ &        3 &        3 &  (35, 35, 25) &  (30, 30, 30) &  (0.38, 0.37, 0.23) &  (0.54, 0.54, 0.53) \\
37 &    3 &    1 &    7 &   30 &  $1.4{\cdot}10^{-13}$ &  $1.5{\cdot}10^{-13}$ &        0 &        0 &  (38, 38, 30) &  (34, 35, 20) &  (0.37, 0.37, 0.26) &  (0.52, 0.52, 0.23) \\
38 &    4 &    1 &    7 &   35 &  $9.2{\cdot}10^{-12}$ &  $5.5{\cdot}10^{-11}$ &        0 &        0 &  (40, 40, 30) &  (38, 38, 28) &  (0.36, 0.36, 0.25) &  (0.51, 0.51, 0.39) \\
39 &    2 &    1 &    3 &    8 &   $2.3{\cdot}10^{-4}$ &   $3.4{\cdot}10^{-4}$ &        3 &        3 &  (35, 35, 35) &  (30, 30, 10) &  (0.38, 0.38, 0.38) &   (0.54, 0.53, 0.1) \\
40 &    3 &    2 &    4 &   15 &   $1.9{\cdot}10^{-9}$ &     $5{\cdot}10^{-9}$ &        2 &        1 &  (38, 38, 38) &  (34, 34, 24) &  (0.37, 0.37, 0.37) &  (0.52, 0.52, 0.27) \\
41 &    4 &    0 &    3 &    8 &   $2.9{\cdot}10^{-7}$ &   $2.2{\cdot}10^{-6}$ &        3 &        3 &  (40, 40, 41) &  (38, 38, 27) &  (0.36, 0.36, 0.36) &  (0.51, 0.51, 0.46) \\
42 &    6 &    1 &    7 &   22 &  $1.2{\cdot}10^{-14}$ &  $6.5{\cdot}10^{-14}$ &        0 &        0 &  (43, 43, 44) &  (42, 42, 32) &  (0.35, 0.35, 0.35) &  (0.49, 0.49, 0.36) \\
43 &    5 &    3 &   13 &   63 &  $1.3{\cdot}10^{-25}$ &  $1.3{\cdot}10^{-24}$ &        0 &        0 &  (42, 42, 32) &  (40, 30, 30) &  (0.35, 0.35, 0.26) &   (0.5, 0.29, 0.33) \\
44 &    4 &    0 &    3 &   17 &   $2.9{\cdot}10^{-7}$ &   $2.2{\cdot}10^{-6}$ &        3 &        3 &  (41, 41, 41) &  (37, 38, 37) &  (0.36, 0.36, 0.36) &  (0.51, 0.51, 0.51) \\
45 &    3 &    0 &    2 &    8 &   $3.4{\cdot}10^{-5}$ &     $2{\cdot}10^{-4}$ &        3 &        3 &  (38, 28, 38) &  (34, 34, 35) &  (0.37, 0.27, 0.37) &  (0.52, 0.52, 0.52) \\
46 &    4 &    2 &    5 &   17 &  $1.6{\cdot}10^{-11}$ &  $5.4{\cdot}10^{-11}$ &        0 &        0 &  (41, 40, 20) &  (38, 27, 38) &  (0.36, 0.36, 0.15) &  (0.51, 0.35, 0.51) \\
47 &    5 &    0 &    4 &   16 &                   $0$ &                   $0$ &        0 &        0 &  (42, 42, 32) &  (40, 40, 40) &  (0.35, 0.35, 0.27) &     (0.5, 0.5, 0.5) \\
48 &    3 &    0 &    3 &   17 &   $4.9{\cdot}10^{-7}$ &   $1.2{\cdot}10^{-6}$ &        3 &        2 &  (38, 38, 20) &  (35, 34, 35) &  (0.37, 0.37, 0.18) &  (0.52, 0.52, 0.52) \\
49 &    3 &    1 &    5 &   24 &  $1.2{\cdot}10^{-14}$ &  $9.9{\cdot}10^{-14}$ &        0 &        0 &  (38, 38, 38) &  (35, 35, 35) &  (0.37, 0.37, 0.37) &  (0.52, 0.52, 0.52) \\
50 &    4 &    0 &    7 &   39 &  $8.3{\cdot}10^{-20}$ &  $6.2{\cdot}10^{-18}$ &        0 &        0 &  (41, 31, 41) &  (37, 38, 38) &  (0.36, 0.31, 0.36) &  (0.51, 0.51, 0.51) \\
51 &    5 &    0 &    4 &   17 &  $1.6{\cdot}10^{-11}$ &  $6.9{\cdot}10^{-10}$ &        0 &        0 &  (42, 42, 42) &  (40, 30, 40) &  (0.35, 0.35, 0.35) &     (0.5, 0.4, 0.5) \\
52 &    2 &    1 &    3 &   15 &   $9.2{\cdot}10^{-6}$ &   $7.7{\cdot}10^{-6}$ &        3 &        3 &  (35, 35, 35) &  (30, 30, 30) &  (0.37, 0.37, 0.38) &  (0.53, 0.53, 0.53) \\
53 &    3 &    1 &    4 &   18 &   $3.9{\cdot}10^{-8}$ &   $1.9{\cdot}10^{-7}$ &        3 &        3 &  (38, 38, 38) &  (35, 35, 34) &  (0.37, 0.37, 0.37) &  (0.52, 0.52, 0.52) \\
54 &    4 &    1 &    7 &   36 &  $6.6{\cdot}10^{-12}$ &  $1.3{\cdot}10^{-11}$ &        0 &        0 &  (41, 41, 40) &  (37, 38, 37) &  (0.36, 0.36, 0.36) &  (0.51, 0.51, 0.51) \\
55 &    6 &    1 &    9 &   43 &  $3.9{\cdot}10^{-26}$ &  $1.1{\cdot}10^{-24}$ &        0 &        0 &  (43, 43, 40) &  (42, 42, 42) &  (0.35, 0.35, 0.33) &  (0.49, 0.49, 0.49) \\
56 &    4 &    1 &    9 &   47 &  $3.8{\cdot}10^{-17}$ &  $1.1{\cdot}10^{-16}$ &        0 &        0 &  (40, 30, 40) &  (38, 37, 38) &   (0.36, 0.3, 0.36) &  (0.51, 0.51, 0.51) \\
57 &    3 &    1 &    6 &   25 &   $7.3{\cdot}10^{-8}$ &   $1.8{\cdot}10^{-7}$ &        3 &        3 &  (38, 30, 38) &  (34, 34, 35) &  (0.37, 0.28, 0.37) &  (0.52, 0.52, 0.52) \\
58 &    2 &    1 &    2 &   15 &   $9.2{\cdot}10^{-6}$ &   $1.7{\cdot}10^{-5}$ &        3 &        3 &  (35, 35, 35) &  (31, 30, 30) &  (0.37, 0.37, 0.38) &  (0.54, 0.54, 0.54) \\
59 &    3 &    1 &    4 &   27 &   $2.4{\cdot}10^{-8}$ &   $3.7{\cdot}10^{-8}$ &        1 &        2 &  (38, 38, 38) &  (35, 35, 34) &  (0.37, 0.37, 0.37) &  (0.52, 0.52, 0.52) \\
60 &    5 &    0 &    4 &   30 &  $4.2{\cdot}10^{-15}$ &  $4.3{\cdot}10^{-13}$ &        0 &        0 &  (42, 42, 42) &  (40, 40, 40) &  (0.35, 0.35, 0.35) &     (0.5, 0.5, 0.5) \\
61 &    6 &    1 &    7 &   32 &            $10^{-16}$ &    $7{\cdot}10^{-16}$ &        0 &        0 &  (44, 44, 44) &  (42, 42, 42) &  (0.35, 0.35, 0.35) &  (0.49, 0.49, 0.49) \\
62 &    5 &    1 &    5 &   24 &  $2.4{\cdot}10^{-12}$ &  $1.4{\cdot}10^{-11}$ &        0 &        0 &  (42, 42, 32) &  (40, 40, 40) &  (0.35, 0.35, 0.23) &     (0.5, 0.5, 0.5) \\
63 &    4 &    2 &    7 &   34 &  $6.4{\cdot}10^{-15}$ &  $7.9{\cdot}10^{-14}$ &        0 &        0 &  (41, 41, 40) &  (37, 37, 38) &  (0.36, 0.36, 0.36) &  (0.51, 0.51, 0.51) \\
64 &    2 &    3 &    6 &   25 &  $1.2{\cdot}10^{-11}$ &  $9.9{\cdot}10^{-12}$ &        0 &        0 &  (35, 35, 35) &  (30, 30, 30) &  (0.37, 0.38, 0.38) &  (0.53, 0.53, 0.53) \\
65 &    4 &    1 &    7 &   27 &  $6.6{\cdot}10^{-12}$ &  $7.7{\cdot}10^{-11}$ &        0 &        0 &  (38, 38, 38) &  (36, 36, 36) &  (0.26, 0.26, 0.26) &  (0.35, 0.35, 0.35) \\
66 &    3 &    1 &    4 &   25 &  $1.7{\cdot}10^{-11}$ &  $3.8{\cdot}10^{-11}$ &        0 &        0 &  (38, 38, 38) &  (35, 34, 34) &  (0.37, 0.37, 0.37) &  (0.52, 0.52, 0.52) \\
67 &    3 &    1 &    5 &   27 &  $4.5{\cdot}10^{-13}$ &  $1.9{\cdot}10^{-12}$ &        0 &        0 &  (38, 38, 38) &  (34, 35, 34) &  (0.37, 0.37, 0.37) &  (0.52, 0.52, 0.52) \\
68 &    3 &    1 &    4 &   17 &   $3.9{\cdot}10^{-8}$ &     $7{\cdot}10^{-8}$ &        3 &        2 &  (38, 38, 38) &  (35, 34, 34) &  (0.37, 0.37, 0.37) &  (0.52, 0.52, 0.52) \\
69 &    3 &    1 &    4 &   18 &   $3.9{\cdot}10^{-8}$ &     $7{\cdot}10^{-8}$ &        3 &        2 &  (38, 38, 38) &  (34, 34, 35) &  (0.37, 0.37, 0.37) &  (0.52, 0.52, 0.52) \\
70 &    5 &    2 &    7 &   26 &  $1.5{\cdot}10^{-15}$ &  $2.4{\cdot}10^{-14}$ &        0 &        0 &  (43, 42, 42) &  (40, 40, 40) &  (0.35, 0.35, 0.35) &     (0.5, 0.5, 0.5) \\
71 &    3 &    1 &    4 &   30 &  $4.5{\cdot}10^{-13}$ &  $1.3{\cdot}10^{-12}$ &        0 &        0 &  (38, 38, 38) &  (34, 35, 34) &  (0.37, 0.37, 0.37) &  (0.52, 0.52, 0.52) \\
72 &    3 &    0 &    2 &   15 &   $9.1{\cdot}10^{-7}$ &             $10^{-5}$ &        3 &        3 &  (38, 38, 30) &  (35, 35, 25) &  (0.37, 0.37, 0.22) &  (0.52, 0.52, 0.27) \\
73 &    2 &    0 &    1 &   13 &   $1.2{\cdot}10^{-4}$ &   $6.8{\cdot}10^{-4}$ &        3 &        3 &  (35, 35, 35) &  (30, 30, 30) &  (0.38, 0.37, 0.38) &  (0.54, 0.53, 0.54) \\
74 &    3 &    1 &    4 &   12 &   $2.7{\cdot}10^{-6}$ &   $5.2{\cdot}10^{-6}$ &        3 &        3 &  (38, 38, 38) &  (34, 34, 35) &  (0.37, 0.37, 0.37) &  (0.52, 0.52, 0.52) \\
75 &    3 &    1 &    3 &    7 &   $2.7{\cdot}10^{-6}$ &   $5.2{\cdot}10^{-6}$ &        3 &        3 &  (38, 38, 38) &  (34, 24, 35) &  (0.37, 0.37, 0.37) &  (0.52, 0.28, 0.52) \\
76 &    4 &    1 &    4 &   12 &   $2.3{\cdot}10^{-8}$ &   $5.6{\cdot}10^{-8}$ &        3 &        3 &  (40, 41, 40) &  (38, 28, 28) &  (0.36, 0.36, 0.36) &   (0.51, 0.26, 0.4) \\
77 &    3 &    1 &    4 &   12 &   $2.7{\cdot}10^{-6}$ &   $3.4{\cdot}10^{-6}$ &        3 &        3 &  (38, 38, 38) &  (35, 35, 35) &  (0.37, 0.37, 0.37) &  (0.52, 0.52, 0.52) \\
78 &    5 &    1 &    6 &   22 &  $1.5{\cdot}10^{-10}$ &   $1.1{\cdot}10^{-9}$ &        0 &        1 &  (42, 42, 42) &  (40, 40, 40) &  (0.35, 0.35, 0.35) &     (0.5, 0.5, 0.5) \\
79 &    5 &    1 &   11 &   42 &  $1.8{\cdot}10^{-23}$ &  $1.6{\cdot}10^{-22}$ &        0 &        0 &  (42, 42, 42) &  (40, 40, 40) &  (0.35, 0.35, 0.35) &     (0.5, 0.5, 0.5) \\
80 &    5 &    0 &    5 &   28 &                   $0$ &                   $0$ &        0 &        0 &  (42, 42, 42) &  (40, 40, 40) &  (0.35, 0.35, 0.35) &     (0.5, 0.5, 0.5) \\
81 &    8 &    0 &   10 &   46 &  $1.6{\cdot}10^{-26}$ &  $1.2{\cdot}10^{-23}$ &        0 &        0 &  (45, 46, 46) &  (44, 45, 45) &  (0.34, 0.34, 0.34) &  (0.47, 0.47, 0.47) \\
82 &    3 &    1 &    3 &   13 &   $2.4{\cdot}10^{-8}$ &   $8.3{\cdot}10^{-8}$ &        3 &        3 &  (38, 38, 38) &  (34, 25, 35) &  (0.37, 0.37, 0.37) &  (0.52, 0.37, 0.52) \\
83 &    4 &    0 &    3 &    7 &   $2.9{\cdot}10^{-7}$ &   $2.2{\cdot}10^{-6}$ &        2 &        3 &  (41, 40, 41) &  (38, 38, 38) &  (0.36, 0.36, 0.36) &  (0.51, 0.51, 0.51) \\
84 &    2 &    2 &    3 &   15 &   $1.6{\cdot}10^{-7}$ &   $6.3{\cdot}10^{-8}$ &        3 &        3 &  (35, 35, 20) &  (30, 30, 30) &  (0.37, 0.37, 0.14) &  (0.54, 0.54, 0.53) \\
85 &    4 &    3 &    8 &   34 &  $1.8{\cdot}10^{-14}$ &  $8.3{\cdot}10^{-15}$ &        0 &        0 &  (41, 41, 41) &  (38, 38, 38) &  (0.36, 0.36, 0.36) &  (0.51, 0.51, 0.51) \\
86 &    4 &    3 &   11 &   49 &  $7.3{\cdot}10^{-18}$ &  $8.1{\cdot}10^{-18}$ &        0 &        0 &  (40, 40, 40) &  (38, 38, 38) &  (0.36, 0.36, 0.36) &  (0.51, 0.51, 0.51) \\
87 &    3 &    1 &    5 &   18 &   $2.7{\cdot}10^{-6}$ &   $5.2{\cdot}10^{-6}$ &        3 &        3 &  (38, 38, 38) &  (34, 35, 34) &  (0.37, 0.37, 0.37) &  (0.52, 0.52, 0.52) \\
88 &    3 &    2 &    5 &   24 &  $8.6{\cdot}10^{-16}$ &  $1.4{\cdot}10^{-15}$ &        0 &        0 &  (38, 38, 38) &  (35, 35, 34) &  (0.37, 0.37, 0.37) &  (0.52, 0.52, 0.52) \\
89 &    6 &    5 &   14 &   74 &  $1.4{\cdot}10^{-51}$ &  $1.1{\cdot}10^{-47}$ &        0 &        0 &  (44, 43, 43) &  (42, 42, 42) &  (0.35, 0.35, 0.35) &  (0.49, 0.49, 0.49) \\
90 &    4 &    0 &    4 &   27 &  $1.2{\cdot}10^{-14}$ &  $1.5{\cdot}10^{-13}$ &        0 &        0 &  (40, 40, 40) &  (38, 37, 38) &  (0.36, 0.36, 0.36) &  (0.51, 0.51, 0.51) \\
91 &    2 &    1 &    3 &   12 &   $2.3{\cdot}10^{-4}$ &   $2.3{\cdot}10^{-4}$ &        3 &        3 &  (35, 35, 35) &  (31, 30, 30) &  (0.37, 0.37, 0.37) &  (0.53, 0.53, 0.53) \\
92 &    4 &    1 &    6 &   23 &  $4.6{\cdot}10^{-10}$ &   $3.2{\cdot}10^{-9}$ &        0 &        0 &  (40, 41, 41) &  (38, 38, 38) &  (0.36, 0.36, 0.36) &  (0.51, 0.51, 0.51) \\
93 &    5 &    1 &    6 &   26 &  $1.9{\cdot}10^{-14}$ &  $1.6{\cdot}10^{-13}$ &        0 &        0 &  (42, 42, 32) &  (39, 40, 40) &  (0.35, 0.35, 0.22) &     (0.5, 0.5, 0.5) \\
94 &    3 &    2 &    5 &   16 &  $2.7{\cdot}10^{-11}$ &  $6.7{\cdot}10^{-11}$ &        0 &        0 &  (38, 38, 38) &  (35, 25, 35) &  (0.37, 0.37, 0.37) &  (0.52, 0.26, 0.52) \\
95 &    3 &    1 &    5 &   26 &   $1.9{\cdot}10^{-9}$ &     $6{\cdot}10^{-9}$ &        2 &        2 &  (38, 38, 38) &  (35, 35, 34) &  (0.37, 0.37, 0.37) &  (0.52, 0.52, 0.52) \\
96 &    3 &    1 &    4 &   16 &   $2.7{\cdot}10^{-6}$ &   $3.4{\cdot}10^{-6}$ &        3 &        3 &  (38, 38, 38) &  (35, 34, 35) &  (0.37, 0.37, 0.37) &  (0.52, 0.52, 0.52) \\
97 &    3 &    1 &    4 &   24 &  $3.4{\cdot}10^{-10}$ &  $7.5{\cdot}10^{-10}$ &        2 &        0 &  (38, 38, 38) &  (34, 34, 35) &  (0.37, 0.37, 0.37) &  (0.52, 0.52, 0.52) \\
98 &    5 &    1 &   11 &   52 &  $2.5{\cdot}10^{-21}$ &  $1.7{\cdot}10^{-19}$ &        0 &        0 &  (42, 42, 42) &  (40, 40, 40) &  (0.35, 0.35, 0.35) &     (0.5, 0.5, 0.5) \\
99 &    4 &    1 &    4 &   18 &   $2.3{\cdot}10^{-8}$ &   $8.3{\cdot}10^{-8}$ &        1 &        3 &  (30, 41, 41) &  (37, 37, 37) &  (0.32, 0.36, 0.36) &  (0.51, 0.51, 0.51) \\
\end{longtable}

\end{landscape}

    

\end{document}